# TENSE, ASPECT AND MOOD BASED EVENT EXTRACTION FOR SITUATION ANALYSIS AND CRISIS MANAGEMENT

A THESIS SUBMITTED TO
THE GRADUATE SCHOOL OF INFORMATICS
OF
THE MIDDLE EAST TECHNICAL UNIVERSITY

BY

ALİ HÜRRİYETOĞLU

IN PARTIAL FULFILLMENT OF THE REQUIREMENTS FOR THE DEGREE OF
MASTER OF SCIENCE
IN
THE DEPARTMENT OF COGNITIVE SCIENCE

April 2012

# TENSE, ASPECT AND MOOD BASED EVENT EXTRACTION FOR SITUATION ANALYSIS AND CRISIS MANAGEMENT

Submitted by Ali Hürriyetoğlu in partial fulfillment of the requirements for the degree of Master of Science in Cognitive Science, Middle East Technical University by,

Prof. Dr. Nazife Baykal                     _______________________

Director, **Informatics Institute**

Prof. Dr. Cem Bozşahin                     _______________________

Head of Department, **Cognitive Science**

Dr. Ceyhan Temürcü                     _______________________

Supervisor, **Cognitive Science, METU**

**Examining Committee Members:**

Prof. Dr. Nihan Kesim Çiçekli                     _______________________
Computer Engineering, METU

Dr. Ceyhan Temürcü                     _______________________
Cognitive Science, METU

Dr. Ruket Çakıcı                     _______________________
Computer Engineering, METU

Assist. Prof. Dr. Bilal Kırkıcı                     _______________________
Foreign Language Education, METU

Assist. Prof. Dr. Murat Perit Çakır                     _______________________
Cognitive Science, METU

**Date:**         <u>27.04.2012</u>

I hereby declare that all information in this document has been obtained and presented in accordance with academic rules and ethical conduct. I also declare that, as required by these rules and conduct, I have fully cited and referenced all material and results that are not original to this work.

Name, Last name:    Ali Hürriyetoğlu

Signature:

_________________

# ABSTRACT

TENSE, ASPECT AND MOOD BASED EVENT EXTRACTION FOR
SITUATION ANALYSIS AND CRISIS MANAGEMENT

Hürriyetoğlu,Ali

M. Sc., Department of Cognitive Science

Supervisor: Dr. Ceyhan Temürcü

May 2012,  95 pages

Nowadays event extraction systems mainly deal with a relatively small amount of information about temporal and modal qualifications of situations, primarily processing assertive sentences in the past tense. However, systems with a wider coverage of tense, aspect and mood can provide better analyses and can be used in a wider range of text analysis applications. This thesis develops such a system for Turkish language. This is accomplished by extending Open Source Information Mining and Analysis (OPTIMA) research group's event extraction software, by implementing appropriate extensions in the semantic representation format, by adding a partial grammar which improves the TAM (Tense, Aspect and Mood) marker, adverb analysis and matching functions of ExPRESS, and by constructing an appropriate lexicon in the standard of CORLEONE. These extensions are based on the theory of anchoring relations (Temürcü, 2007, 2011) which is a cross-linguistically applicable semantic framework for analyzing tense, aspect and mood related categories. The result is a system which can, in addition to extracting basic event structures, classify sentences given in news reports according to their temporal, modal and volitional/illocutionary values. Although the focus is on news reports of natural disasters, disease outbreaks and man-made disasters in Turkish language, the approach can be adapted to other languages, domains and genres. This event extraction and classification system, with further developments, can provide a basis for automated browsing systems for preventing environmental and humanitarian risk.

# ÖZ

## DURUM ANALİZİ VE KRİZ YÖNETİMİ AMAÇLI ZAMAN, GÖRÜNÜŞ VE KİP EKLERİ TABANLI OLAY ÇIKARIMI

Hürriyetoğlu, Ali

Yüksek Lisans, Bilişsel Bilimler

Tez Yöneticisi: Dr. Ceyhan Temürcü

Mayıs 2012, 95 sayfa

Günümüzde olay çıkarımı sistemleri temel olarak görece az miktarda bir bilgiyi, durumların zaman ve kipsel niteliklerinden, öncelikle geçmiş zamanı niteleyen bildirimleri işlemektedir. Fakat zaman, görünüş ve kip eklerini inceleyebilen sistemler daha iyi analizler sağlayabilir ve daha geniş uygulama alanı bulabilirler. Bu proje, Türkçe için bu tarz bir sistem geliştirmektedir. Bu sistem, Open Source Information Mining and Analysis (OPTIMA) araştırma grubunun olay çıkarımı yazılımlarına anlamsal gösterim formatı kapsamında gerekli eklemeler yapmıştır. Bu eklemeler, kısmi gramerlerin zaman, görünüş ve istek ekleri için uyarlanması, zarfların incelenmesi ve ExPRESS eşleme fonksiyonlarının geliştirilmesidir. Aynı zamanda, CORLEONE yazılımı standardında sözlük hazırlanmıştır. Bu iyileştirmeler, bütün dillerde uygulanabilecek, zaman, görünüş ve istek ilişkili kategoriler içeren bir anlamsal çerçeve olan çapa ilişkileri kuramına (Temürcü, 2007, 2011) göre yapılmıştır. Çalışmanın sonucunda geliştirilen sistem, basit olay yapısının çıkarımına ek olarak, haber kaynaklarındaki cümleleri, zamansal, kipsel ve isteksel değerlerine göre sınıflandırır. Her ne kadar analiz Türkçe dilinde doğal afet, hastalık salgınları ve insan kaynaklı kaza haber raporlarına odaklansa da, bu yaklaşım diğer dillere, alanlara ve yazı türlerine uyarlanabilir. Bu olay çıkarımı ve sınıflandırma sistemi, ek geliştirmelerle, çevresel ve insani zararların engellenmesi için otomatik tarayıcı sistemlerin geliştirilmesini sağlayabilir.

**Anahtar Kelimeler**: Çapa İlişkileri Kuramı, Olay Çıkarımı, Kriz Yönetimi, Durum Analizi, Haber Analizi

# ACKNOWLEDGMENTS

I would like to thank my supervisor Dr. Ceyhan Temürcü and to Vanni Zavarella for their invaluable support, suggestions, guidance and patience throughout this study. Without their support, this thesis could not have been completed.

This thesis project started during my one-year traineeship in European Commission Joint Research Centre (Ispra, Italy). I am indepted to the Open Text Information Mining and Analysis Group (OPTIMA) for the stimulating working environment and to open-hearted friends there.

I want to express my gratitude to Erik van der Goot and Dr. Ralf Steinberger for their continuous advise and support in every phase of my study.

I would like to thank Vanni Zavarella (again), Hristo Tanev and Marco Turchi for sharing their knowledge and experience in event extraction systems with me. They provide me the basis to start and pursue this study.

I would like to thank Prof. Dr. Yaşar Özden for accepting me as a research assistant at the Computer Technology and Instructional Technology Department. Thanks to all the knowledge he and department's faculty shared with me during my assistantship there.

This work also owes to the support of my family and many life-long friends. And mostly to Hatice Dodak, because we accomplish it together from the very beginning.

To my parents

# TABLE OF CONTENTS

# CHAPTER 1

# 1. INTRODUCTION

The thesis project aims to develop an event extraction software system which can classify news content into combinations of semantic TAM categories developed in Temürcü (2007, 2011). This system is intended to provide analysts a tool for browsing and searching texts for temporal and modal values in a fast and effective way. The system is tested with an aggregated corpus of news articles on natural disasters, man made disasters and disease outbreaks. As a prospect, the system is expected to provide practical data for people or agent systems who are in charge of planning urgent initiatives in crisis situations like natural disasters, man-made disasters, or epidemic diseases.

Although the current project focuses on extracting TAM-related information from disaster texts in Turkish language, the developed system should be adaptable to any other language and other genres by preparing related lexicons and modifying grammar rules.

Human languages refer not only to entities, but critically, also to situations. Therefore, various aspects of situations are worth analyzing in modeling linguistic meaning. Formal languages which attempt to model natural languages will provide better basis for theories of mental representation and reasoning only if they take into account various aspects of language use, including the situatedness of sentences with respect to temporal, epistemic and volitional reference points.

This situatedness of  human linguistic messages in time, in knowledge frames and in intentions should be considered as an essential point in any semantic representation format designing process. Accordingly, this study aims to convert natural language sentences into an expressive and compact semantic representation format which captures temporal, epistemic, and volitional modalities. This is achieved by adding new features to flat feature structures used in Open Source Information Mining and Analysis (OPTIMA) research group's event extraction methods (Tanev, Piskorski, & Atkinson, 2008). By implementing necessary extensions to the system – namely; constructing a Turkish lexicon compatible with the extended semantic format in the standard of CORLEONE which is a set of loosely coupled, general purpose basic lightweight linguistic processing resources, and writing a partial grammar which builds on the capabilities of the extraction pattern engine ExPRESS – a system will be implemented which converts Turkish sentences into the intended semantic representation.

This extended representation format, which takes into account not only the propositional contents of human messages, but also their temporal, epistemic and volitional/illocutionary values, can in principle be used in computational models of high-level cognitive processes, including situation awareness, reasoning, and planning.

A system of event extraction sensitive to aforementioned values of sentences should ideally exploit all kinds of linguistic and non-linguistic information on temporal constituency, temporal location, epistemic modality, evidentiality, as well as the desires and illocutionary intentions of speakers/writers. Although in spoken language these information sources can be supplemented by prosodic or even visual cues, we aim to analyze only textual information and try to detect and process some lexical and grammatical clues which signal temporal, epistemic and volitional qualifications of a situation.

In addition to grammatical tense, aspect and mood markers (TAM), adverbs and other lexical cues may also help anchor a situation in a particular temporal, epistemic or volitional frame. Therefore, our analysis will also use some adverbial cues in the preparation of lexical resources and in the adaptation of grammatical processor.

In the theory of anchoring relations developed in Temürcü (2007), every finite sentence is analyzed as expressing one value in each of the temporal, epistemic and volitional domains. In other words, every sentence can be mapped to a "TAM class", consisting of a combination of three anchoring categories (temporal, epistemic, volitional).

Steps for creating a computational system that accomplishes this classification is explained in the following chapters, after a literature review on related topics. The system uses a manually created lexicon, which consists of  morphological variations of verbs and adverbs frequently encountered in the news genre, as well as  nominal and verbal expressions which are of interest in disaster/crisis contexts. Since at present there is no study on frequencies of occurrence of such words in Turkish, a limited corpus was used and the relevant words were identified manually. Although machine learning techniques could have been used for automatizing this task, this would require further work which would be beyond the immediate scope of the intended study.

# CHAPTER 2

# 2. LITERATURE REVIEW

## 2.1 A representational format for temporal and modal values of situations

The classical basis for computational models of semantics of human languages is the first order logic (FOL) formalism. FOL has been applied to various linguistic phenomena such as event structure (Dowty, 1989), thematic roles (Parsons, 1990) , tense (Kamp & Reyle, 1993) , modalities by accounting for possible worlds, plurals (Link, 1983) or anaphoric expressions (Eijck, 2005; Kamp & Reyle, 1993).

The state of the art in the use of FOL in modeling linguistic semantics includes wide-coverage semantic analysis tools such as DORIS (Bos, 2001) and Boxer (Bos, 2008). DORIS translates English texts into Discourse Representation Structures (DRS) (Kamp, 1981), dealing with a wide range of linguistic phenomena including scope ambiguities, pronoun resolution, presupposition, nominal specifications and modality, using lambda-calculus. On the other hand Boxer uses Combinatory Categorial Grammar (CCG) (Steedman, 2001) to build DRSs. The outputs of both systems can be converted to ordinary first-order logic formulas, which can be used in inference tasks. A comparable general aim system for Turkish language is TOY (Say, Demir, Çetinoğlu, & Öğün, 2004). Although TOY is not as comprehensive as Boxer, it translates Turkish sentences to predicate logic formulas and to Prolog statements in a knowledge base. TOY can be considered as a useful interface to various applications like conversational agents, natural language interface, speaking assistants etc. in Turkish language.

Knowledge representation systems which stem from computer science typically have a broader coverage of conceptual relations, while often lacking lexical and grammatical components. Examples of these systems include KL-ONE (Brachman, 1978) and CYC (Lenat & Guha, 1990). When supplemented by parsers and interpreters, such systems can also be used for (specific or wide-coverage) conceptual analyses of natural languages. For example, MultiNets (Helbig, 2010) is an extended knowledge representation framework, coupled with the WOCADI framework, which makes syntactic and semantic analysis of the German language. As a whole, this system is used in various natural language applications (question answering, dialog systems, database interface systems, etc.).

The TimeML specification language mainly deals with event and temporal expressions in natural language texts. It can systematically anchor event predicates to various temporal expressions, order event expressions in a text, and enable the interpretation of underspecified or partially given temporal expressions.

Although the aforementioned formalisms and systems provide general ways for representing/computing propositional contents and support efficient reasoning mechanisms, they lack in certain capabilities. DORIS and Boxer can handle a wide-range of natural language phenomena, but they lack explicit representational mechanisms for modal qualifications, including epistemic modality, evidentiality and volitional moods. MultiNets+WOCADI can recognize, parse and translate questions, but cannot handle other sentence types and modalities, and ignore tense and aspect. The TimeML format is expressive for temporal events, but does not consider epistemic and volitional aspects of language use. In fact, to our knowledge, none of the representational formalisms and computational systems available today can afford a full analysis of temporal, epistemic and volitional/illocutionary aspects of sentences.

Although these extension can in principle implemented in FOL or in Description Logics, which have better computational properties as to decidability (Bos 2011), they do not always satisfy practical needs, like representing TAM classes and provide easy identification of crucial information in crisis situations. What is practically needed is a balance between representation capabilities and processing. The extraction and labeling of relevant information can provide adequately compact and expressive outputs for both human readers and machines. Indeed, information extraction can be considered as a response to the low performance of general language analysis/understanding systems.

This thesis project aims to detect and extract clues of temporal, epistemic and volitional aspects of natural language fragments by a top-down, shallow parsing method, using a cascaded grammars method. Sentences will be classified into TAM classes, which can be considered as comprising a predefined ontology.

## 2.2  Information and Event Extraction Systems

In early phases of computational treatment of collections of texts, detection of relevant documents was seen to be an adequate goal. However, the fact that information on the Web today is often unstructured necessitates 'information extraction' (Appelt, 1999) and 'event extraction' (Cunningham, 2005; Grishman, 2010) systems, which can process contents of documents and seek out relevant information.

The extraction of relevant events enable people to reach to specific events, analyze and track them easily. The current fields of application of these methods include security, management of disease outbreaks, medicine, bioinformatics, analyses of blogs, reports and news data (Grishman, Huttunen, & Yangarber, 2002a, 2002b; Ji & Grishman, 2008; King & Lowe, 2003; Naughton, Kushmerick, & Carthy, 2006; Yangarber, 2005). Tanev et al. (2009) show the multilingual applicability and usability of such event extraction systems.

In order to have an acceptable performance, such extraction systems need to consider a trade-off between complexity and specificity of extracted information: The more complex the data to be extracted, the more specific must be the domain of discourse; the simpler the data, the more generally the extraction algorithms may be applied (Cunningham, 2005).

Event-extraction studies have been motivated by Message Understanding Conferences (MUC) (Grishman & Sundheim, 1996; MUC, 1991, 1992, 1995, 1998; B. Sundheim, 1993; B. M. Sundheim, 1991) and after them, by the Automatic Content Extraction (ACE) (ACE, 2005; Doddington et al., 2004) program. MUC program defined and evaluated five tasks. These are:

(a) Named Entity Recognition (NER) for finding and classifying names of people, places, organizations, dates, amounts of money, etc.,
(b) Coreference Resolution (CR) for identification of identity relations between entities discovered in NER or pronouns,
(c) Template Element (TE) construction in order to add descriptive information to NER results
(d) Template Relation (TR) construction which finds a small number of possible

relations between entities, like an employee relationship between a person and a company, a family relationship between two persons, etc.,

(e) Scenario Template (ST) production discovers TE and TR relations relevant to event descriptions.

Since MUC program devoted to work mainly on entities and their relation to each other and basic predefined events, it does not consider any TAM-related linguistics phenomena. However, it has generally supported and shaped information extraction studies.

The ACE program is a successor to the MUC program. ACE program tasks are:

(a) Entity Detection and Tracking (EDT), Named entity recognition and coreference resolution tasks of MUC program combined for this task,

(b) Relation Detection and Tracking task, is constructed from Template element construction and template relation construction tasks of MUC program,

(c) Event Detection and Characterisation task is a variant of scenario template task. This program takes into account more complex phenomena than the MUC program. It aims to detect all mentions of entities, 24 relation types such as owner, founder, client, part-whole, subtype, located, based in, near, parent, sibling, associate, etc., as well as mentions of events, relations and temporal expressions.

ACE evaluation task includes the detection of modality and tense attributes for relations and events. Polarity and genericity are also attributes of verbs to be extracted. Everything that is evaluated has a type and a subtype, which are part of a predefined ontology.

TempEval is another information extraction task. It uses the TimeBank corpus (Pustejovsky, Hanks, et al., 2003) which is annotated in TimeML (Pustejovsky, Ingria, Setzer, & Katz, 2003) format. The outputs of the system is also in TimeML format. TempEval-1 (Verhagen et al., 2007) and TempEval-2 (Pustejovsky & Verhagen, 2010) tasks focus on events, temporal expressions, and ordering relations among events.

The BioNLP is a shared task on event extraction (Kim, Ohta, Pyysalo, Kano, & Tsujii, 2009) in biological domain. This task uses methods of (a) core-event and primary arguments recognition, (b) extracting secondary arguments, (c) recognition of negation and speculation statements. Due to this last component, this shared task can be considered relevant to epistemic TAM specifications.

Semantic roles-based methods focus on basic event structure on the basis of semantic roles like actor, instrument, affected, etc. (Mccracken, Ozgencil, & Symonenko, 2005). These systems need consistent semantic role label definitions and annotation of verb's predicate/argument structure, such as PropBank (Palmer, Gildea, & Kingsbury, 2005), and FrameNet (Baker, Fillmore, & Lowe, 1998). These systems do not use any tense, aspect or mood markers-based method.

As an instance of rule-based systems, Yakushiji et al. (Yakushiji, Tateisi, Miyao, & Tsujii, 2001) apply a full grammar parsing method to find main arguments of a sentence. Event information is extracted by domain specific mapping rules from argument structures to frame representations. This approach mainly focuses on the predicate and arguments in a sentence, and can not take into account temporal, epistemic or volitional properties.

Another detailed method in event extraction is applied in Fillmore, Narayanan, & Baker, (2006). This system focuses on extracting events on the basis of FrameNet project. Frame bearing words (verbs, nouns, adjectives and prepositions) and grammatical constructions based on these words are detected. This system can in principle capture various phenomena related to TAM classes. However, their focus is on the arguments of the verb, without considering the verb's temporal, aspectual and modal values.

Uzzaman and Allen (2011) apply deep semantic parsing and hand-coded extraction rules to detect events and temporal relations between them. This system detects temporal expressions according TimeML standart and classifies events according an ontology that is similar to FrameNet. It is a robust system in terms of temporal expressions and event detection However, although its ontology classifies verbs in a detailed manner, temporal, epistemic and volitional distinctions are not explicitly given.

Pattern based methods search for certain types of events and their occurrence patterns through the application of a dedicated cascaded grammar (Tanev et al., 2009; Tunaoglu et al., 2009). Patterns are identified according some relevant ontology. Although, to our knowledge, these systems are not used for detecting different TAM class-related features, adding required patterns and grammar rules can enable them to extract and classify sentences according to their TAM values.

In sum, the aforementioned event extraction systems all deal with propositional contents of events. Only some tasks, especially those based on TimeML, also extract certain temporal dimensions. Bioinformatic studies additionally try to include probabilities, speculations and beliefs as much as possible (Vincze, Szarvas, Mora, Ohta, & Farkas, 2010). However, none of these systems consider the full spectrum of temporal, epistemic and volitional aspects of sentences. In addition to temporally aware systems, some studies begin to seek methods for detection of future references to events in Internet (Baeza-yates, 2006). However, these methods are still not comprehensive enough to be considered as full extraction systems.

Every event extraction system requires some sort of ontology since it needs to identify what kind of information should be extracted and represented. The coverage of that ontology limits the scope of the system. An ontology may consider any domain, like conflict or disease outbreak (Kawazoe, Chanlekha, Shigematsu, & Collier, 2008; Lee, 2003). These can also be coupled to multilingual synonym sets in order to work for several languages at the same time.

The current thesis project applies a pattern based method, and uses a TAM ontology inspired by cognitive linguistics studies. Although the focus is on TAM categories and not on event participants, it is important to extract them too, for a complete representation of events. To this end, the current project performs person and group detection. Due to its crisis management scope, it also detects words which refer to various types of disasters, verbs which express requests, and nouns which express humanitarian needs.

### 2.3 Event Extraction Systems in Crisis Management

Early warning and real-time information gathering for helping crisis management is a developing research field. The cumulated information can be analyzed in order to reach critical information in a fast and effective way.

A crisis is any event that may exert harm to the environment and to humans. These events could be e.g., disease outbreaks, man-made disasters, and natural disasters. The severity of these types of events leads us build systems that enable us to understand the situation as fast as possible.

Analyses can include real-time monitoring of important information resources, simple information retrieval, machine translation, automatic summarization, event extraction and so on. The current study focus on extraction of relevant information on the basis of an ontology which provides cues on disasters, requests, and humanitarian needs. The system is supplemented by a TAM ontology which captures temporal, epistemic and volitional aspects of events. Clearly, such TAM-related information is valuable in assessing critical information in contexts of crisis.

The identification of the source, reliability and temporal features of events can be crucial in crisis situations. The result of such a system can enable humans and machines to focus on particular aspects of the given information. The vast amount of information produced by internet technologies can be analyzed effectively according to dimensions provided by a TAM ontology.

Some potential information resources for an event extraction system for crisis management could be the Web sites of humanitarian aid organizations (e.g. Turkish Red Crescent; Kızılay), of disaster monitoring systems (e.g. Kandilli Observatory and Earthquake Research Institute), or public information exchange platforms like Facebook or Twitter.

The recently happened earthquake in Van province of Turkey showed us how people shared critical information on the social media. Many users provide information about different aspects of the disaster. Specific tags were used to identify relevant information. Even official bodies or responsible organizations have spread relevant information on Twitter. Social media analysis has the advantage to be fast and detailed. We can reach information without waiting any article to be published in news.

Although the current thesis project analyzes news articles, it can be adapted to recruit information from social media or any other information resource.

# CHAPTER 3

# 3. TAM CATEGORIES

The task aims to analyze Turkish inflectional TAM markers and adverbials in terms of the semantic framework of anchoring categories of Temürcü (2004, 2007, 2011). TAM markers are analyzed in three domains; which are the temporal, the epistemic and the volitional domains. According to Temürcü (2007) every sentence can be mapped to a value in each of these three domains. Below are some examples of such mappings (with descriptive category labels):

(1) Over the years, people who haven't been vaccinated are now giving the virus a big opportunity to spread.[1]

      ("present", "certainty", "assertion")

(2) Preventive measures will minimize damage from natural disasters.[2]

      ("future", "certainty", "assertion")

---

1  http://www.foxnews.com/health/2011/12/02/measles-outbreaks-on-rise-across-europe/ , 20.12.2011

2  http://india.blogs.nytimes.com/2011/12/19/an-urbanizing-india-faces-natural-disaster-risk/ , 20.12.2011

(3) Help anti-whaling![3]

        ("future", "epistemic possibility", "prescription")

The correspondence of  sentences to combinations of categories in these domains can be used to build a semantic classification on the basis of TAM distinctions. These "anchoring categories" are explained in next parts.

### 3.1 Temporal Anchoring Categories

The time of an event, relative to a reference time, can be identified as fitting in one of the seven categories presented below. Reference time is typically the reporting time in news reports.

These categories can in principle be expresses by grammatical markers, lexical items like temporal adverbs, or by a combination of them (Temürcü, 2007).

      **Table 1.** Temporal anchoring categories. From Temürcü (2007: 55).

| |
|---|
| SIMULTANEOUS presents an event as unfolding or a state as obtaining at the temporal center. (i.e., it expresses an 'immediate condition'.) |
| ANTERIOR presents a SoA as having occurred before the temporal center. |
| POSTERIOR presents a SoA as potentially occurring after the temporal center. |
| PERFECT expresses a past SoA as well as its effects at the temporal center (immediate conditions). |
| PROSPECTIVE expresses a potential event in the future as well as its causes at the temporal center (immediate conditions). |
| RECURRENT expresses a repeated occurrence of a state-of-affairs (SoA) in a temporal range around the temporal center. It can specify a frequency of occurrence which may range from 'rarely' to 'always'. |

---

3  http://www.southasiamail.com/news.php?id=102381 , 20.12.2011

| |
|---|
| ATEMPORAL presents the occurrence of a SoA as free from any temporal restriction relative to the temporal center. |

## 3.2 Epistemic Anchoring Categories.

The epistemic (including evidential) qualifications of information is critical in assessing implications. The classification of epistemic categories in Temürcü (2007) takes into account the relation of the expressed proposition to a reference knowledge state, yielding the following seven categories:

**Table 2.** Epistemic anchoring categories. From Temürcü (2007: 70).

| |
|---|
| NEW INFORMATION presents a proposition as new and not well-integrated into the current knowledge state (i.e., it expresses a piece of immediate evidence.) |
| CERTAIN presents a proposition as already well-assimilated to the current knowledge state. |
| HYPOTHETICAL presents a proposition as unknown relative to the current knowledge state. |
| INFERRED expresses a definitive (certain) conclusion on the basis of immediate evidence. |
| CONJECTURED expresses a non-definitive (uncertain) conclusion on the basis of immediate evidence. |
| PROBABLE expresses an evaluation as to the likelihood of the truth of a proposition. It can specify a degree of probability which may range from 'very unlikely' to 'virtually certain'. |
| GENERAL FACT presents a proposition as one for which the truth is taken as generally valid, rather than as being part of a personal knowledge base. |

### 3.3 Volitional Anchoring Categories

In the theory of anchoring relations (Temürcü 2007, 2011), illocutionary force derives from volitional qualifications of utterances. Speaker's intention and volitional attitude towards the expressed proposition are made explicit by seven volitional anchoring categories:

**Table 3.** Volitional anchoring categories. From Temürcü (2007: 90).

| |
|---|
| IMMEDIATE CONTRIBUTION expresses a reflection which stems from the speaker's will in a specific communicative context. In other words, it directly conveys the speaker's immediate (illocutionary) intentions |
| ACCEPTED presents a reflection as one which is acknowledged to be real. |
| ENVISIONED presents a reflection as one which is in the realm of imagination, rather than as one which is part of the accepted reality. |
| ASSERTED makes an intentional contribution to the communicative context by expressing a reflection that the speaker takes as real. |
| WANTED makes an intentional contribution to the communicative context by expressing the speaker's desire for the realization of an envisioned reflection. |
| ACCEPTABLE presents a reflection as consented (or, conceded) to some degree rather than as fully accepted. |
| GENERAL STATEMENT voices a general announcement of an impersonal (social or institutional) authority, rather than making a personal assertion tied to the will of an individual in a specific communicative context. |

# CHAPTER 4

# 4. PRELIMINARY CONSIDERATIONS IN ANALYZING THE NEWS CONTENT

Event reports convey information on the time of events, their reporting time, and sometimes the identity and the information source of the reporter. Identification of crucial information is important for analyzing the situation rapidly. The following example demonstrates a piece of text in a news report:

(4)

> Van Belediye Başkanı Bekir KAYA: "Şu anda kent merkezinde 3 noktada göçükler meydana geldi. Ancak bu panik ortamında çok fazla bilgi kirliliği var. Doğal olarak ilk deprem anında her kurum ne yapabilirim paniğiyle bir şeyler yapmaya çalıştı. Yardım ekipleri yolda ama hala pek ulaşan olmadı bölgeye. Ben tüm yardım kuruluşlarına şunu söylemek istiyorum. Lütfen hiç talep beklemeden herkes elinden gelen yardımı hemen ulaştırmaya baksın. Kimse Van'dan talep beklemesin. Herkes yardım için elinden geleni yapsın." "Su, ekmek ve çadıra ihtiyacımız var!"

"En çabuk ulaşılacak yerler Erciş ve yakınındaki köylerdir. Şu anda çadırlara, battaniyeye, içme suyuna, temel gıda maddelerine, suya, ekmeğe çok ama çok acil ihtiyaç var."

English translation:

Mayor of Van city, Bekir KAYA: "By now, there are three sites with collapsed buildings in the city center. However, there is too much information pollution in this panic atmosphere. Naturally, every institution tried to do something after the first earthquake. Aid teams are under way, however none of them has reached to the region. My message to all humanitarian organizations: Please try to deliver the aid immediately, as much as you can, without waiting for any request. Do not wait for any request from Van. Everybody should do as much as they can."

"We need water, bread and tents."

"Places that can be reached quickly are Erciş and the villages around it. At the moment, there is an urgent need for tents, blankets, food, water and bread. "

This example is an excerpt from the 14-1-VanDeprem[4] file of the created corpus. It is an example of an early report in a time of crisis; it explains what has happened in the Van province area due to earthquake and mentions required humanitarian aid, e.g. water, bread, blankets, and tents.

Consequent reports of a disaster would consist of updated information from a wider perspective, further details about probable and reasonable plans, emerging needs and what has been done up to current time. The analysis of that information can show what had been done and should be done for every step. That information flow is worth analyzing in order to understand and manage the crisis.

---

4  http://www.haberturk.com/yasam/haber/681925-su-ekmek-ve-cadira-ihtiyacimiz-var-  , 24.10.2011

Analyzing clusters of related news reports provide more complete information than a single document based approach. The current system can be adapted to use document clusters in order to gather critical information across documents. In such a setting, event information collected from chronologically structured reports can be browsed anytime for any TAM category. For example one can search which information is hearsay, which one is certain or which information point to near future.

Using TAM classification enables a system to identify volitional, epistemic and temporal values of a sentence. That method can provide an effective information organization in an automated setting. One important challenge in an automated assignment of sentences into TAM classes is that of ambiguity, because TAM markers and marker combinations, especially grammatical ones, are known to be highly polysemous (Temürcü, 2007, 2011). This can be seen in the following examples, where the "past" marker *-DI* is used for anterior (simple past) as well as for perfects, and the "continuous aspect" marker *-Iyor* is used for simultaneous (present continuous), persisting perfect, and atemporal (generic statements).

(5)

  *(anterior)*
  Kızılay, yardımda bulunmak isteyenlerin 168'i arayabileceğini de **açıkladı**.
  'Turkish Red Crescent **stated** that people who would like to help should dial
  168.'

  *(perfect of result)*
  Erçiş'te 25 apartman ve bir öğrenci yurdu **çöktü**.
  'In Erciş 25 buildings and a student dormitory have **collapsed**.'

  *(perfect of recent past)*
  Bölgedeki kan merkezleri de yaralılar için gerekli kanı ulaştırmak için çalışmalarına **başladı.**
'Blood centers in the region have **started** to work in order to provide required blood for wounded people.'

(simultaneous)
Tesislerde Ankara'dan gelen uzman ekip incelemelerde **bulunuyor**.
'An expert team from Ankara is analyzing the situation at the facility.'

*(persisting perfect)*
Üç gündür bölgeden haber **alınamıyor**.
'There **have been** no news from the region for three days.'

*(atemporal)*
Bu siyanürün çok küçük miktarı bile **zehirliyor**.
'Even a very small amount of cyanide **is poisonous**.'

Lexical cues, mainly adverbs and adverbial phrases, can be used to reduce such ambiguities, although some ambiguity may remain even after analyzing adverbs.

(6)
　　**Bugün sabah** Kızılay tarafından hazırlanan 11 ton ağırlığında kahvaltılık malzeme gönderildi.
　　**This morning**, 11 tons of breakfast food, prepared by Turkish Red Crescent, was sent to the region.
　　"Bugün sabah" (*this morning*) makes it clear that the temporal category involved is ANTERIOR

　　ATB'nin yardım tırı **yarın sabah** yola çıkıyor.
　　ATB's aid truck leaves **tomorrow morning**.

"Yarın sabah" (*tomorrow morning*) makes it clear that the temporal category involved is POSTERIOR

Olaya tepki gösteren bölge halkı ile tesisin özel güvenlikleri arasında **zaman zaman** gerginlik yaşandı.

There were **occasional** quarrels  between the local community that react against the incidence and private security guards of the facility

"Zaman zaman" (*occasionally, from time to time*) makes it clear that we have REPETITIVE in the scope of ANTERIOR.

**Şu anda** insanı zehirleyen miktarın 125 katı bir kirlilik söz konusu.

**At this moment**, the level of pollution is 125 times more that the minimum level for human poisining

"Şu anda" (*at this moment*) makes it clear that the temporal catregory is SIMULTANEOUS, rather than, say, ATEMPORAL.

Köylülerin **belki de** bu hayvanları yememesi gerekiyor.

**Maybe**, villagers should not eat these animals.

"Belki de" (*maybe*) makes it clear that the epistemic catergory is PROBABLE, rathar than, say, CERTAIN.

**Edinilen bilgilere göre** kurtarma ekibi bölgeye henüz ulaşmadı.

**According to reports**, the rescue team has not reached to the region yet.

"Edinilen bilgilere göre" (*according to reports*) makes it clear that the volitional (illocutionary) category involves HEARSAY as taking scope over ASSERTION.

Text in the news domain is relatively restricted in terms of lexicon and grammatical structures (Bell, 1996; Semino, 2009). It is hence possible to use sublanguage grammars and lexicons. A sublanguage is a particular language fragment used to communicate some common goals or discuss particular topics. Sublanguage Theory studies (Liddy, Jorgensen, Sibert, & Yu, 1991) have shown that various types of texts such as news, reports, email, manuals, requests, arguments, interviews etc. differ in terms of lexical, semantic, discourse and pragmatic features (Mccracken et al., 2005).

Lexicon preparation process in the current project takes a sublanguage approach. Adding just highly occurring structures to the lexicon can be considered as implicit annotation of the corpus. In that approach the corpus size is critical, since a small corpus would not cover important structures. On the other hand, a very big corpus would include unrelated structures. Therefore we try to add most frequent and important lexical structures for aforementioned domains e.g. disease outbreak, natural disasters, man-made disasters. With a sublanguage approach, by adapting the lexicon and the grammar rules, TAM categories in news reports can be analyzed with a higher precision than in general verbal communication.

One of the most frequent TAM marker in Turkish news reports is -*DI* (and its enclitic counterpart -*(y)DI*). -*DI* typically expresses ASSERTED as a volitional dimension, CERTAIN as an epistemic dimensions and PERECT or ANTERIOR as temporal dimensions. It can express epistemic distance and counter-factuality too. -*(y)DI* typically states various kind of anaphoric past (ANTERIOR_ana). This values may take scope over ATEMPORAL, SIMULTANEOUS or PERFECT_persisting, depending on the combinations with other inflectional TAM markers (-*mIŞ, -AcAk, -Iyor, -mAktA*, etc.).  You can find some usage examples of -*DI* from Van earthquake 14-2-VanDeprem[5] file below.

---

(7)

a.

Başkanlığımızdan 20 kişilik bir teknik heyet Van'a intikal et**ti**. Ayrıca, çeşitli illerden 200 teknik personel görevlendiril**di**.

A technical committee of 20 people were transferr**ed** to Van from our presidency. Additionally, 200 technical staff from various provinces were authorized.

b)

3 adet askeri kargo uçağı ile personel ve malzeme transferi yapmak üzere, Ankara-Van arasında ulaşım zinciri oluşturul**du**.

A chain of transportation **has been** creat**ed** between Ankara and Van to transport personnel, materials and equipments by 3 military cargo aircrafts.

Although the crucial part of an article is the first paragraph, details are generally found in the body of an article. After explaining what has happened briefly at the beginning, following parts consist of details about probable aspects of an event, future directions or inferred information. The system developed in this project analyzes the whole of a news report in order to get as much as information possible.

Turkish time expressions and temporal structure can provide a basis for the analysis of the temporal domain (Şeker & Diri, 2010; Yusuf Karagöl, 2010). Furthermore the document time can be considered as the current time of analysis in online news event extraction systems. Therefore indexical use of time adverbials can be disambiguated to some extent. The issue of how other discourse-related considerations will help reduce possible ambiguities in TAM class assignments to Turkish sentences is left as a future study. Some prospects on this point will be mentioned in the conclusion.

# CHAPTER 5

# 5. COMPONENTS AND SUBTASKS OF THE SYSTEM

The developed software is a converter from Turkish natural language to an extended semantic representation format, which is a feature structure. This structure consists of TAM-related features as well as some interesting features of NPs, e.g. name, and quantifier.

The system output consists of feature-value pairs for recognized items in analyzed sentences. This output can be seen as a mapping from natural language sentences to predefined formats. This mapping is mediated by an ontology of TAM categories. The conversion is mainly based on the main verb of the sentence, as well as adverbial expressions.

Textual analysis and identification of the sentence structure is done with OPTIMA Action's in house software, which is written in Java programming language. The software consists of two main parts, which are Corleone (Piskorski, 2008) for tokenization, sentence splitting and other low level processes and ExPRESS (Piskorski, 2007), for application of grammatical rules in order to detect values for sentence elements which are identified in the lexicon part. The extensions in the current project have been implemented in Java programming language.

An important advantage of these systems is the possibility of extending them by defining new features, grammars and grammar application functions for particular purposes. The current study has improved the system for verb and adverb based analyses.

The input of the system is news articles which are divided into paragraphs and sentences. The aim is to be able to recognize TAM category bearing structures, e.g. verbal forms, adverbials, which are encoded in the lexicon. Since the sentences that occur in news have a limited grammatical variation, this approach categorizes the structures well enough for purposes of situation analysis and crisis management.

This conversion tool can be in use for different tasks as well, e.g. as a pre-annotator for an annotation gold standard for TAM-based sentence categorization.

## 5.1 Linguistic Structures Analysis

A full semantic analysis of a natural language fragment necessitates solutions for complex phenomena, including syntactic scope, disambiguation, pronoun resolution, quantifier analysis, and so on. Today there is no single deterministic and effective solution for these issues. Instead of taking all these phenomena into account, the system developed in this project converts sentences to feature structures with predefined concepts. These concepts are defined in the lexicon and grammar files. In this way, sentences are effectively categorized into TAM classes, through the detection of TAM-marked verb forms and adverbials.

The mapping of the natural language to that predefined TAM ontology is guided by a partial grammar. Verbs are divided into different categories according to their grammatical and semantic properties. Other elements that support the verb, including adverbials and noun phrases, are also detected in order to extract important event information. Categorization of verbs and other syntactic categories is explained in the lexicon preparation section 5.4.

The project is based on verb forms since the verb is the pivotal element in the event structure. Adverbs have been taken into account in the grammar in order to disambiguate the TAM categories and add additional information to the event structure. Reported speech verbs (dedi, söyledi, aktardı) are treated as volitional/illocutionary modifiers of the main verb (T. L. Hobbs, Karmarkar, & Hobbs, 1997). Assigning relevant values to volitional feature enables the system to take 'hearsay' information into account.

The grammar is applied to each line of an article sequentially until the end of the current file. A line may consists of a paragraph or a sentence. Allowing paragraphs to be processed as a whole allows the system to use discursive clues and leave space for using any additional mechanism that takes into account longer fragments than a sentence e.g. discourse connectives, and coordinators. One of these discursive clues is that sequential sentences tend to have same or similar TAM categories. This point has been left for further research (see Chapter 7).

## 5.2 Mapping to TAM Categories

The occurrence of a category dimension bearing term is used in the same way as in the studies of opinion mining and sentiment analysis (Pang & Lee, 2008). Category bearing terms are detected if they occur in the lexicon. When different categories are denoted by more than one expression, dedicated grammar rules give priority or combine these according to dedicated implemented procedures. These procedures are explained in next parts. Although sentiment analysis studies mainly use statistical methods (Malinsk, 2010), category bearing terms are more structural and predictable by grammar rules (Tan et al., 2011).

Although there are several high precision morphological analysis tools for Turkish language (Eryigit & Adali, 2004; Oflazer, Göçmen, & Bozşahin, 1994; Sak, Güngör, & Saraçlar, 2010; Say et al., 2004; Çöltekin, 2010), adding the verb's and other elements' morphological variations to lexicon with their TAM values is the most effective option for news domain (and probably, for other domains as well). A look up to lexicon may be more efficient than analyzing morphology. Since the verbal forms, with their suffixes, are most of the times unique, recognition of verb forms by grammar rules decrease the ambiguity in TAM category assignment process.

Once we have an ontology for TAM categories, the generation of morphological variations of verbs and their TAM values can in principle be done automatically. But since this study is exploratory in this field, verb forms and their TAM values have been recorded to the lexicon manually.

Creation of an article-based and cluster-based collection of news and prepare them as a core corpus was the first step of this study. Preparing a list of verb forms, adverbials and reported speech verbs which can facilitate the assignment of sentences into TAM classes was the second step.

TAM classes were taken from a subset of the TAM assignments tables of Temürcü (2007: 175-177), on the basis of some frequently occurring TAM combinations in news reports (Appendix A).

Preparation of a document-based corpus and construction of the lexicon are explained in the following sections.

### 5.3 Corpus Preparation

The corpus was gathered manually from on-line news articles. New articles were added when they were in a related domain: natural disasters, man-made disasters and disease outbreaks.

The numbering of the articles is topic based. Every unique topic has a number. The articles which are about same topic are numbered under the same number such as 1.2, 1.3, according their report time.

This method of corpus organization allows topics to be extended just by adding new topics by giving the consequent number. Every topic can be considered as a stack since numbering is done according to publishing date of articles. Therefore the corpus can be considered as a set of stacks.

The type of the article is encoded as MMD for man-made disasters, ND for natural disasters and DO for disease outbreaks. Those abbreviations are used in the name of the file which consists the article.

Example :

    1-KutahyaSiyanur-MMD

        1-1-KutahyaSiyanur-MMD

      …

    2-KarsHeyelan-ND

    3-KaradenizTunaNehri-MMD

    ...

This example shows only first files in the corpus. The first topic (KutahyaSiyanur-MMD) is a cluster that has at least two articles. The second topic has just one article.

Numbering documents chronologically inside the clusters can help track the temporal ordering of events related to a certain topic. Although out of the scope of this study, taking this ordering into account together with the TAM analyses can provide a basis for analyzing causal and epistemic relations among events.

The corpus is stored in text files, and each article contains its publication date, URL and author.

## 5.4 Lexicon Preparation

The system reads lexicon entries and analyzes the target text by using these entries. Since the ExPRESS system (Piskorski, 2007) processes flat feature structures, entries in the lexicon are prepared as flat feature structures. Flat feature structures enable the system to be fast and be able to perform several levels of shallow parsing.

The current entries of the lexicon are first names, quantifiers, numerals, person group bearing words, titles, post modifiers, clitics, verbs and adverbials. Verbs have been divided into several categories that are explained below. Any TAM category bearing element can be added when it is required for further or better analysis. Every entry has the related feature-value pairs in order to be used in matching grammars.

Feature values of verb forms which have TAM markers are taken from the well-defined tables in Temürcü (2007: 175-177) (see Appendix A for some TAM combinations frequently encountered in news reports). Since some markers map to more than one TAM class, which is a reflection of natural polysemy, the result of a look up to lexicon can be ambiguous. In order to decrease this ambiguity, instead of adding all the variations from the table for a marker, we simplified the reference table to include only frequently occurring TAM classes in the news genre. This approach enables us to add as less as possible variations for each entry. That solution also lets us exclude infrequently occurring feature-value combinations for some verbs forms. Although not adding all variations for each verb form can be considered as an incompleteness, a middle size corpus can be created just as comprehensive as required. Even one superfluous lexical entry can make the disambiguation process very complex and lead to wrong results, since adverbial and other clues may not always be sufficient for disambiguation. That would cause problems in a discourse-centric approach, since the false disambiguations may cause cumulative errors.

Using the corpus to identify lexical units that should be added to the lexicon can be considered as implicitly annotating that corpus. However, this approach was taken here only to control the representativeness of the lexicon. In addition to extraction from corpus, some frequent entries can be identified by machine learning techniques (Tanev et al., 2009), which can be applied on random news articles from the Web.

Next section explains how lexical entries and their values are defined for the system.

### 5.4.1 Lexicon Entry Features and Feature Values

The lexicon is prepared in the standard format of CORLEONE (Core Linguistic Entity Online Extraction) (Piskorski, 2008). This is a meta-engine which provides a basis for defining features and their values. It has several modules for low-level linguistics processing: (a) a basic tokenizer, (b) a tokenizer which performs fine-grained token classification, (c) a component for performing morphological analysis, (d) a memory efficient database-like look-up component, and (e) a sentence splitter.

The compilation of a lexicon by CORLEONE produces a file in a finite-state automaton format. The ExPRESS system takes this compiled file and grammar rules as input, in order to analyze a text.

The feature structures accepted by the system are completely user-defined. Features are specified in the configuration files. The values of the feature structures are user defined as well. However, values are not specified in the configuration files, since the system interprets values as it reads them from the lexicon file and uses them in grammatical rules. The grammar uses these feature-value pairs and sequence of elements to detect required patterns.

The first part of an entry is the entry word which should be found in the tokenized input file. The type of the word  is encoded in the GTYPE feature which used in the grammar to denote which type of words should be matched. These types are -defined according to requirements of particular tasks.

Although all of the features of an entry can be used in finding matched items according to grammar rules, in our settings GTYPE is used as a starting point to let just predefined types of entries to be taken into account.

Another general feature is SURFACE, which is used for defining the output string. This feature can be arranged to normalize the output. The value of that feature can be used directly or manipulated by additional functions during the grammar application to get the required output string. Although the current task uses the value of SURFACE feature as the form as the entry word, it can be changed as other requirements emerge, e.g. into the infinitive form or lemma form.

In the following examples, values of GTYPE are gaz_given_name for first names, gaz_initial for initials, turkish_numeral for number bearing words, turkish_numeral_post_modifier for modifiers which occur after numbers, gaz_quantifier_req_number for quantifiers which denotes the numbers after them and gaz_title is for titles. Meaning of the entries in English is given just after the entry in the parenthesis.

Example:

      Ergin | GTYPE:gaz_given_name | SURFACE:Ergin

      Ufuk | GTYPE:gaz_given_name | SURFACE:Ufuk

      A | GTYPE:gaz_initial | SURFACE:A

      K | GTYPE:gaz_initial | SURFACE:K

      hiç | GTYPE:gaz_neg_quantifier | SURFACE:hiç (any)

      hiçbir | GTYPE:gaz_neg_quantifier | SURFACE:hiçbir (nothing)

      bir | GTYPE:turkish_numeral | SURFACE:bir (a)

      iki | GTYPE:turkish_numeral | SURFACE:iki (two)

      onlarca | GTYPE:turkish_numeral | SURFACE:onlarca (dozens)

      civarında | GTYPE:turkish_numeral_post_modifier | SURFACE:civarında (around)

      kadar  | GTYPE:turkish_numeral_post_modifier | SURFACE:kadar (up to)

      takriben | GTYPE:gaz_quantifier_req_number | SURFACE:takriben (around)

      yaklaşık | GTYPE:gaz_quantifier_req_number | SURFACE:yaklaşık (around)

      phd | GTYPE:gaz_title | SURFACE:phd

      prof. dr. | GTYPE:gaz_title | SURFACE:prof. dr.

      bekar | GTYPE: person_modifier | SURFACE:bekar  (bachelor)

      emekli | GTYPE: person_modifier | SURFACE:emekli (retired)

Named entity and person group recognition procedures use these entry types in order to identify prominent actors or victims of events. Moreover, detection of these elements can be followed by a search for adverbs around them. This increases the coverage of the analysis in terms of searched space.

During the tokenization process, CORLEONE software separates the words and their suffixes  by apostrophe. These suffixes are most of the time case markers. Therefore these case markers have been encoded in the lexicon in order to be able to recognize them after apostrophe structures. Thus, the grammar application process can recognize and handle these structures by using dedicated grammar rules.

Example:

      ye | GTYPE: case_dative | SURFACE:ye (dative case marker)

      i | GTYPE: case_accusative | SURFACE:i (accusative case marker)

      de | GTYPE:turkish_clitic | SURFACE:de (turkish clitic)

GNUMBER feature is used for coding plurality or singularity of the entry. That mechanism is used in detecting number agreement.

Example:

      Asyalı | GTYPE:person_group_proper_noun | GNUMBER:sg | SURFACE:Asyalı (from Asia)

      Orta Doğulular | GTYPE:person_group_proper_noun | GNUMBER:pl | SURFACE:Orta Doğulu (from middle east)

      Aktris | GTYPE:person_position | GNUMBER:pl | SURFACE:Aktris (actress)

      Aktörler | GTYPE:person_position | GNUMBER:pl | SURFACE:Aktör (actor)

In the example SURFACE feature is normalized to be in singular form of the entry.

Quantifiers comprise another important syntactic category encoded in the lexicon. This entry type enables the system to recognize internal structures of noun phrases.

Example:

      bazı | GTYPE:gaz_quantifier | SURFACE:bazı (some)

      epey | GTYPE:gaz_quantifier | SURFACE:epey (quite)

TAM classes are encoded as combinations of VOLITIONAL, EPISTEMIC and TEMPORAL features. Their values define their position in each dimension. The verb type is encoded in the GTYPE feature:

Example:

> geldi | GTYPE:report_verb | VOLITIONAL:asserted | EPISTEMIC:certain | TEMPORAL:anterior | SURFACE:geldi ("has come", "came", ...)

> bulunuyor | GTYPE:verb | VOLITIONAL:asserted | EPISTEMIC:certain | TEMPORAL:simultaneous | SURFACE:bulunuyor (be present, ..)

TAM values of sentences can have specifications which describe finer-grained distinctions than those provided by anchoring categories. These can be specifications are provided by adverbs, or by grammatical markers which may denote, e.g., specific types of PERFECT (such as perfect of result or perfect of persisting situation). Examples include 'galiba' (*apparently*) and 'belki' (*perhaps*) in the epistemic domain; 'dün' (*yesterday*), 'hep' (*all the time*), 'zaman zaman' (*from time to time*) in the temporal domain. Since temporal specifications are widely used, these are represented in TEMPSPEC feature in the feature structures of verbs and sentences. However, specifications for other (epistemic and volitional) TAM dimensions are not represented due to their less frequent occurrence:

Important verb categories include report verbs, evidential verbs and epistemic possibility verbs. Since these categories are used inclusively in news articles, proper treatment of these type of verbs improve the system's coverage. These verb forms, such as 'söyledi' (*said*) or 'anlaşıldı' (*it turned out that*) signal the epistemic nature of propositions. Generally these provide additional information about the modality of the event and the reliability of the information (T. L. Hobbs et al., 1997). Report verbs, evidential verbs and possibility verbs which do not directly denote an event are encoded in the lexicon in a different way than normal verbs. According to Hobbs et al. (1997), this strategy facilitates focusing on the main verb of the sentence. The report_verb, evidential_verb and possibility_verb values of GTYPE feature implement such a strategy.

Examples:

> belirtti | GTYPE:report_verb | VOLITIONAL:hearsay | SURFACE:belirtti ("stated", "has stated")

> anlaşıldı | GTYPE:evidential_verb | VOLITIONAL:asserted | EPISTEMIC:inferred | SURFACE:anlaşıldı ("turned out")

> ihtimali var | GTYPE:possibility_verb | VOLITIONAL:asserted | EPISTEMIC:probable | SURFACE:ihtimali var ("is likely")

Each subordinate verb category has its relevant TAM features defined such as HEARSAY for reporting verbs, INFERRED or CONJECTURED for evidential verbs and PROBABLE for epistemic possibility verbs.

Relativized and passive forms of report and evidential verbs have also been added, since these forms are used frequently.

Example:

> belirten | GTYPE:report_verb | VOLITIONAL:hearsay | SURFACE:belirten (who indicates)

düşünülen | GTYPE:evidential_verb | VOLITIONAL:asserted | EPISTEMIC:conjectured | SURFACE:düşünülen (which is considered )

tahmin edilen | GTYPE:evidential_verb | VOLITIONAL:asserted | EPISTEMIC:conjectured | SURFACE:tahmin edilen (which is estimated)

Since the verb form is different in main clauses than in subordinate contexts, both forms are entered in the lexicon. In the following example, the verb in (a) is in normal form, while in (b), in the subordinate verb form:

Example:
a) Van'a iş çevrelerinden yardım **yağıyor**!
Aid is **pouring** from the business community to Van.
b) Almanya Sağlık Bakanlığı ise salgında vaka sayısının **azaldığını** duyurdu.
German Ministry of Health announced **a decrease** in the number of disease cases.

There are three different subordinate verb forms encoded in the lexicon: Factive subordination, future subordination and action subordination. Markers for the factive subordination are *-DIGI/-DIGINI/-DIGInA/-DIGIndAn* as in (a) in the following example, markers of future subordination are *-AcAGI/-AcAGInI/AcAGInA/AcAGInDAn* as in (b), and action subordination markers are *-mAsI/-mAsInI/-mAsInA/-mAsInDAn*, as in (c):

Example:
a)
zorunda kaldığı | GTYPE:subordinate_verb | TEMPORAL:simultaneous | SURFACE:zorunda kaldığı (was obliged to)

zorunda kaldığına | GTYPE:subordinate_verb |
TEMPORAL:anterior_ana_occ | SURFACE:zorunda kaldığına

zorunda kaldığını | GTYPE:subordinate_verb |
TEMPORAL:anterior_ana_occ | SURFACE:zorunda kaldığını

b)

neden olabileceği | GTYPE:subordinate_verb | TEMPORAL:prospective |
SURFACE:neden olabileceği (may cause)

neden olabileceğine | GTYPE:subordinate_verb | TEMPORAL:posterior |
SURFACE:neden olabileceğine

neden olabileceğine | GTYPE:subordinate_verb | TEMPORAL:prospective |
SURFACE:neden olabileceğine

neden olabileceğini | GTYPE:subordinate_verb | TEMPORAL:posterior |
SURFACE:neden olabileceğini

c)

direniş görmesi | GTYPE:subordinate_verb | TEMPORAL:prospective |
SURFACE:direniş görmesi (encounter resistance)

direniş görmesini | GTYPE:subordinate_verb | TEMPORAL:prospective |
SURFACE:direniş görmesini

direniş görmesine | GTYPE:subordinate_verb | TEMPORAL:posterior |
SURFACE:direniş görmesine

TAM values of subordinate verb forms are only specified for the temporal feature.
Other feature values are completed during the grammar application phase, according to the subordinating verb form.

Another verb category includes the verbs that denote some event or situation happening as auxiliary verbs. Since these verbs need another event or nominalizations of event names they have their particular method of evaluation. These verbs and the nouns that occur before these verbs are given in the lexicon as shown in the following examples. Some of these nouns have the GTYPE value 'disaster', since they describe disaster-related situations.

Example:

a)

meydana geldi | GTYPE:incidence_verb | VOLITIONAL:asserted | EPISTEMIC:certain | TEMPORAL:anterior_ana_occ | SURFACE:meydana geldi (happened)

söz konusudur | GTYPE:incidence_verb | VOLITIONAL:asserted | EPISTEMIC:general_fact | TEMPORAL:atemporal | SURFACE:söz konusudur (it is the case that)

b)

depreminin | GTYPE:disaster | SURFACE:deprem (earthquake)

fırtına | GTYPE:disaster | SURFACE:fırtına (storm)

Some specific meanings of verbs are also considered in order to capture important situations. Since the chosen news domain includes humanitarian needs reported in case of emergencies, the lexicon includes a special class for verbs which denote 'requests', and a special class for NPs which denote 'humanitarian needs'. Examples of such entries are below.

Example:

a)

gerekiyor | GTYPE:request_verb | VOLITIONAL:asserted |
EPISTEMIC:new_information | TEMPORAL:posterior |
SURFACE:gerekiyor (is needed)

ihtiyacı çekiyor | GTYPE:request_verb | VOLITIONAL:asserted |
EPISTEMIC: new_information | TEMPORAL:simultaneous |
SURFACE:ihtiyacı çekiyor (needs)

ihtiyaç var | GTYPE:request_verb | VOLITIONAL:asserted | EPISTEMIC:
new_information | TEMPORAL:simultaneous | SURFACE:ihtiyaç var (there
is a need)

b)

acil müdahale | GTYPE:humanitarian_need | SURFACE:acil müdahale
(emergency intervention)

battaniye | GTYPE:humanitarian_need | SURFACE:battaniye (blanket)

Adverbial expressions constitute another important part of the analysis. The selection
of adverbs for each domain of TAM markers has been done on the basis of previous
studies and the corpus. Since adverbs denote just one dimension of TAM categories,
their definition and is done just according to their related TAM dimension. The
GTYPE values of adverbial expressions are adverb_v, adverb_e, and adverb_t,
standing for volitional, epistemic and illocutionary adverbs, respectively. Grammar
rules about that mechanisms can be found in the grammar application sections.

Examples:

      belki | GTYPE:adverb_e | EPISTEMIC:probable | SURFACE:belki (maybe)

      önceki gün | GTYPE:adverb_t | TEMPORAL:anterior | SURFACE:önceki gün (the day before)

      kesinlikle | GTYPE:adverb_e | EPISTEMIC:certain | SURFACE:kesinlikle (certainly)

      o halde | GTYPE:adverb_e | EPISTEMIC:inferred | SURFACE:o halde (then)

Accusative and possessive forms of person positions, person group names, disaster and humanitarian need names have been entered, since these forms occur frequently. However, these case markers are not marked explicitly by features, since the analysis does not focus on them. The feature values are just the same as they were in nominative case:

Example:

      müzisyenleri | GTYPE:person_position | GNUMBER:pl | SURFACE:müzisyenleri (musicians)

      müşavirleri | GTYPE:person_position | GNUMBER:pl | SURFACE:müşavirleri (consultants)

      Kaçakçıların | GTYPE:person_position | GNUMBER:pl | SURFACE:Kaçakçıların (traffickers)

Kaçakların | GTYPE:person_position | GNUMBER:pl |
SURFACE:Kaçakların

Negative verbs forms are directly added as well. Some lexicon entries for negative verbs are shown below. Although polarity could have been explicitly specified as a feature, since the focus is on TAM values, affirmative and negative forms have been entered in the same way.

Example:

kalır | GTYPE:verb | VOLITIONAL:asserted | EPISTEMIC:general_fact |
TEMPORAL:atemporal | SURFACE:kalır (remains)

kalmaz | GTYPE:verb | VOLITIONAL:asserted | EPISTEMIC:general_fact |
TEMPORAL:atemporal | SURFACE:kalmaz

The next section explains the grammar rules used for performing the intended tasks.

## *5.5 Grammatical Rules*

## 5.5.1 Some Syntactic Considerations

The analysis mainly focuses on detecting the verb form and adverbials in order to identify the TAM class of a sentence.

Although Turkish has the scrambling feature, the flexibility of word order is not infinite. Especially, the official style used in reporting news articles often limits the sentences to the canonical SOV ordering. Similarly, the canonical position of adverbs in Turkish is immediately before the verb (Wilson & Saygın, 2001). However, in practice there is great deal of flexibility. Especially in news articles temporal adverbials tend to occur at the beginning of the sentences. That is why we have two different strategies for producing the output. First strategy is to search for adverb just before the verb and the second way is to look for the adverb further away. For the second option, a set of grammar rules were used to search for adverbials up to three positions before the verb. This is implemented both in verb- and sentence grammar rules.

There seems to be two ways an adverb can effect the TAM class of a sentence. Most adverbs only add a specification to verbs. On the other hand, some adverbs can override the TAM class of the verb, such as adding anaphoric features or restricting possible class assignments.

Since the grammar rules are partial, they do not cover whole sentences as single patterns. They are also applied to sub-sentential verbal clauses. This enables the system to discover more than one event in long sentences that consists of several verbal expressions. Although the grammar is partial, including different verbal forms as subordinating verbs enables the system to analyze almost the whole sentence.

## 5.5.2 Person and Person Group Detection

Several variations of person name or person group mentions are detected by the system. Person names can be combinations of first and last name, their initials and their title information. Person group expressions can also include information about position or number, as shown in the following examples:

Example:

a) Person name:

　　　Mehmet Korkmaz , M. K. , Dr. Mehmet Korkmaz

b) Person group

　　　yüzlerce kişi (hundreds of people), 6 mühendis (6 engineers)

Grammar rules make use of the GTYPE feature with following values for that part of the task; gaz_given_name, gaz_initial, gaz_neg_quantifier, turkish_numeral, turkish_numeral_post_modifier, person_group_proper_noun, person_position, gaz_quantifier, gaz_title, person_modifier, case_ablative, case_dative, case_accusative, case_locative and case-genitive.

First names and initials have central roles in identifying person names. After identifying a first name if we have another simple name or an initial with a dot after it, we can decide that this sequence is mentioning a person name, e.g., Ali Hürriyetoğlu or Ali H. Several variations of these sequences are supported as well.

Another important entry groups are person positions and person group proper nouns. These nouns have quantifiers, numerals and modifiers in order to identify person groups like 'bütün öğrenciler' (*all students*), 'çoğu kişi' (*most people*), '200 kadar öğretmen' (*about 200 teachers*).

Case markers that occurs after an apostrophe are recognized as well. This method enables the system not to loose track of grammar rules especially after numbers like "6'sı mühendis".

Furthermore, after identifying first person and person groups, the system searches whether there is a sequence of these names. That sequence can be separated by a comma or coordination bearing lexical unit such as 've' (*and*), 'ile' (*with*), 'ile birlikte' (*together with*), 'yanısıra' (*along with*).

## 5.5.3 Verb based grammar application

Verbs have been divided into five different categories. These are: (1) general verb category, including request verbs and incidence verbs, (2) report verbs, (3) evidential verbs, (4) possibility verbs, and (5) subordinating verbs.

Request verbs and incident verbs are treated identically with general verbs in terms of TAM classification, since they already have all the TAM features in their feature structures. But report verbs, evidential verbs and possibility verbs combine with subordinating verbs in order to yield a complete TAM class.

### 5.5.4 Calculating the Effects of Adverbs

Adverbial clauses add specific meanings to the TAM values of sentences. Therefore taking into account adverbial expressions enables the system to make more detailed and precise analyses.

Adverbs may occur everywhere in the sentence. However, the analysis focus on the adverbs that are close to verbal structures. Since the canonical order of the adverb is just before the verb in Turkish, most adverbs are straightforwardly detected. The current implementation searches for adverbs up to three places before the verb in the sentence. This search can cover the whole sentence by slightly changing the grammar rules, extending the searched space.

The effect of adverbial clauses is basically modifying a specific TAM category. Since every adverb has just one dimension (temporal, epistemic or volitional), application is done just for one feature for each adverb. Some adverbs only add a special value to the verb's TAM value. For instance, in the following sentence, the expression 'saat beşte' (*at five o'clock*) only adds a specific value to the temporal category of the verb, which is ANTERIOR:

Example:

      Ekip saat beşte ayrıldı.
      *The team left at five o'clock*

In such cases, the TEMPSPEC feature keeps track of temporal specifications provided by adverbs. In case a temporal adverb is matched by a grammar rule, the adverb and it's TAM value are kept in this feature.

A second way in which an adverb may influence the TAM value of a sentence, is completely changing the TAM class normally signaled by a verb form. The verbal form 'kaybedecek' (*will loose*) will normaly denote certainty, but the adverb 'belki de' (*maybe*) denotes probability. In this case, the modal feature of the adverb determines the epistemic status of the whose sentence. After having parsing the output, an appropriate filtering will give us the correct TAM class for the sentence.

> Belkide yüzlerce insan hayatını kaybed-ecek.
> maybe "hundreds of" human "lose one's life"-FUT
> *'Maybe, hundreds of people will loose their lives.'*

A third type of influence of adverbs consists in reducing potential ambiguities stemming from the polysemy of verb forms. For instance, the suffix *-(y)AcAG* in Turkish can be used for both POSTERIOR or PROSPECTIVE. When it occurs with an adverb like 'bundan sonra' (*from now on*), the sentence should analyzed as expressing PROSPECTIVE. Similarly, some temporal adverbs like 'o zaman' (*by that time*) or 'daha sonra' (*then*), may clarify that temporal reference is anaphoric rather than being deictic.

The following example shows an analysis produced by the system. The text fragment "zaman zaman gerginlik yaşandı"(there were occasional quarrels) is recognized as a verbal structure with an adverb occurring before it. The SURFACE feature shows the relevant parts of the text. Other features show the TAM category dimension values. The VOLITIONAL and EPISTEMIC values are taken from the lexicon entry of the verb "gerginlik yaşandı"(there were quarrels). The adverb "zaman zaman" (occasionally, from time to time) has the temporal feature value 'repetitive', and the rule called anaphoric_temp_adv_1 adds this value to the verb's temporal value, which in turn becomes anaphoric (anterior_ana). The calculation hence yields the value "anterior_ana_repetitive":

Example:

      zaman zaman gerginlik yaşandı

      verbal_analysis

      TYPE : Verb

      SURFACE : zaman zaman gerginlik yaşandı

      VOLITIONAL : asserted ( )

      EPISTEMIC : certain ( )

      TEMPORAL : anterior_ana_repetitive ( )

      TEMPSPEC : repetitive \ zaman zaman \

      RULE : anaphoric_temp_adv_1

Another example of calculating the effect of an adverb is shown below. In case an ATEMPORAL adverb like "hiçbir zaman" (*never*) occurs before a verb with -*(y)AcAG*, the resulting verbal structure of the combination will become PROSPECTIVE.

Example:

      hiçbir zaman zararsız olmayacaktır (.. will never be harmless)

      verbal_analysis

      TYPE : Verb

      SURFACE : hiçbir zaman olmayacaktır

      VOLITIONAL : asserted ( )

      EPISTEMIC : general_fact ( )

      TEMPORAL : prospective

TEMPSPEC : atemporal \ hiçbir zaman \
RULE : atemporal_to_prospective_1

The next section explains how all relevant information is combined into a unified sentence feature structure.

In addition to adverb effect calculation, subordinating verbs merge their features with the subordinate verbs for a complete TAM class, as these both have partial representations of TAM values. For example, in case we have a report verb like 'söyledi' (*said*), its TAM category dimension, which is HEARSAY, is combined with those of the subordinate verb. Examples of this calculus will be presented in Section 6.1.

## 5.5.5 Sentence Structure

The system aims to detect TAM dimensions of whole sentences. Therefore, grammar rules were adapted to identify TAM dimensions of sentences, on the basis of verbs and adverbs which occur in the sentence. Since the cascaded grammars in the system focus on verbs, they start to search for other sentence elements starting from verbs. This feature limits the search space and sometimes analyses do not cover all the sentence, especially the beginning of a sentence. But since adverbs are generally close to verbs, most of the cases are analyzed accurately.

The result of the system is a feature structure that contains, (i) the surface form of the identified items in the sentence, (ii) The TAM values of the sentence, (iii) person or person groups under a title called 'people', (iv) and cues for any humanitarian need.

The Following example shows a sentence feature structure.

Example:

      sonraki dönemde ise kansere yol açabilmektedir. ( .. at the next period may cause cancer)

      sentence_analysis

      TYPE : Sentence

      SURFACE : yol açabilmektedir (may cause to occur)

      VOLITIONAL : asserted ( )

      EPISTEMIC : general_fact ( \ \ )

      TEMPORAL : simultaneous ( )

      VERB : yol açabilmektedir

      DISASTER : kanser (cancer)

      RULE : sentence

In this sentence structure not all features have a value; the system did not find any NPs which denote people or humanitarian needs.

## 5.5.6 The Grammar Component

### 5.5.6.1 General Features and Structure of ExPRESS

ExPRESS (Extraction Pattern Recognition Engine and Specification Suite) is a pattern engine that provides a platform for specifying and processing cascaded finite-state grammars (Piskorski, 2007). It is a hybrid of JAPE which is used in GATE system and the unification-based XTDL which is used in SProUT. Grammar rules are regular expressions over feature structures. The system consists of two main parts: a) cascaded grammar interpreter and b) grammar parser. Grammar specification is divided in to three parts: a) type declaration, b) a set of grammar definitions, c) work flow specification.

### 5.5.6.2 Grammar Definitions

Types declaration part is a list of all used types and appropriate features of these types that can be used in the grammars. Main types can be found in the following examples. These types are defined in a different file from grammar rules. A configuration file connects them in order to recognize used features by defining the runtime usage of the grammar rules.

Example:

> gazetteer := [GTYPE,VOLITIONAL,EPISTEMIC,TEMPORAL,SURFACE,
>           GNUMBER, AMOUNT, SLOTTYPE, NONUM, CONTEXT,
>           NONVIOLENT]
>
> person :=  [NAME,  TYPE, FIRST_NAME, LAST_NAME, INITIAL1,
>           INITIAL2, SEX, POSITION,TITLE, RELIABLE, RULE]
>
> person_group := [NAME,TYPE,QUANTIFIER, AMOUNT,RULE]
>
> quantity := [NAME,QUANTIFIER,TYPE,AMOUNT,RULE]
>
> verbal_analysis := [TYPE, SURFACE, VOLITIONAL, EPISTEMIC,
> TEMPORAL, TEMPSPEC,RULE]
>
> adverbial_v := [TYPE,SURFACE,VOLITIONAL]
>
> adverbial_e := [TYPE,SURFACE,EPISTEMIC]
>
> adverbial_t := [TYPE,SURFACE,TEMPORAL]
>
> adverbial_t_ana:=[TYPE,SURFACE,VOLITIONAL,EPISTEMIC,TEMPOR
> AL]
>
> sentence_analysis := [TYPE, SURFACE, VOLITIONAL, EPISTEMIC,
> TEMPORAL, TEMPSPEC, VERB, PEOPLE, HUMANITARIAN_NEED,
> DISASTER, RULE]

The type 'gazetteer' identifies what can be recognized in the text from lexicon entries. It has all possible features in order to be able to be used in grammars by arranging the grammars for specific values of features.

Other types; 'person', 'person_group' and 'quantity' are used in NP recognition. The essential types for TAM category identification are 'verbal_analysis' and adverbials which have TAM category bearing features e.g. VOLITIONAL, EPISTEMIC, TEMPORAL.

Other general features are RULE for storing matched rule name and HUMANITARIAN_NEED for keeping needs of disaster victims.

The grammar definition file starts with a grammar configuration part, which may consists of a list of arbitrary processing resources which are applied before the interpreter applies the grammar. After the resources definition, a search strategy can be chosen among the following options: (i) longest-match, (ii) all-matches and (iii) all-longest-matches.

Example:

        SETTINGS:
        {
          MODULES: <CorleoneTokenizer>, <CorleoneBasicTokenizer>,
        <CorleoneGazetteer>
          SEARCH_MODE: all_longest_matches
          OUTPUT: grammar
        }

This example represents how grammar settings are defined. Used external resources are CorleoneTokenizer, CorleoneBasicTokenizer and CorleoneGazetteer. The search mode is "all-longest-match", in order to have all possible outputs for whole sentences. The OUTPUT part specifies what to pass over to next grammar level. In this case the value is 'grammar', which passes to the next level only patterns for further grammar application. It could also be 'all', which would send to next level whatever is available, such as the outputs of other modules like gazetteer.

Grammar rules are specified between "PATTERNS" and "END_PATTERNS" labels. The left hand side (LHS) part of a rule is a regular expression over flat feature structures (FFS), which are non-recursive typed feature structures (TFS) without coreferencing. Information transport from LHS to the right hand side (RHS) is enabled by string-valued attributes which are tailored variables of LHS. Moreover, functional operators enable slot values of matched features to be manipulated on the RHS in order to produce combined outputs.

The following example illustrates a rule for matching person names. It matches a sequence of items with the type 'gazetteer', which can be outputs of the dictionary look-up tool for 'gaz_given_name', 'gaz_initial', 'case_accusative', 'case_dativ', 'case_genitiv', 'case_ablativ'. Other items are: 'token', which is the output of the tokenizer, and 'basic-token', which is a specified string. The question mark after an item specification shows that it is not mandatory for that item to occur. Additionally, the symbol & denotes a link between a type name of the FFS with its feature-value pairs. This represents the constraints that should be fulfilled, and the symbol # introduces a variable name independently of any binding to the surface forms of the matched text fragments. The label 'name' at the end of LHS marks the start/end position of the defined action on the RHS of the rule. RHS action produces a 'person' structure by accessing variables from LHS and computing result structures with function calls such as 'ConcWithBlanks()' and Conc().

Example:

```
person_name  :> ((gazetteer & [GTYPE: "gaz_given_name", SURFACE: #base_form])

                ((gazetteer & [GTYPE: "gaz_initial", SURFACE: #initial_1] token & [SURFACE: "."])
                   | (gazetteer & [GTYPE: "gaz_initial", SURFACE: #initial_1] token & [SURFACE: "."]
                      gazetteer & [GTYPE: "gaz_initial", SURFACE: #initial_2] token & [SURFACE: "."])) ?

                (gazetteer & [GTYPE: "gaz_name_infix", SURFACE: #infix]) ?

                (token & [TYPE: "first_capital_word", SURFACE: #last_name]
                   | (token & [TYPE: "mixed_word_first_capital", SURFACE: #last_name])
                   | (token & [TYPE: "word_with_hyphen_first_capital", SURFACE: #last_name])
                   | (token & [TYPE: "word_with_apostrophe_first_capital", SURFACE: #last_name]))

                ((basic-token & [SURFACE: "'"] | basic-token & [SURFACE: "'"] | basic-token & [SURFACE: "'"]
                   | basic-token & [SURFACE: "´"]  | basic-token & [SURFACE: "`"] )

                ((gazetteer & [GTYPE: "case_accusative", SURFACE: #case3])
                   | (gazetteer & [GTYPE: "case_dativ", SURFACE: #case3])
                   | (gazetteer & [GTYPE: "case_genitiv", SURFACE: #case3])
                   | (gazetteer & [GTYPE: "case_ablativ", SURFACE: #case3]))) ?

                ((basic-token & [SURFACE:"("])
                 (token & [TYPE: "any_natural_number", SURFACE: #numb1])
                 (basic-token & [SURFACE:")"]) ) ?

                (gazetteer & [GTYPE: "turkish_clitic"]) ? ):name

-> name: person & [NAME: #full_name, TYPE: "N_PER", FIRST_NAME: #base_form, LAST_NAME: #last_name_final, INITIAL1: #initial,RULE: "person_name"]
& #last_name_final :=ConcWithBlanks(#infix, #last_name)
& #initial := PersonNameInitial(#initial_1,#initial_2)
& #full_name := ConcWithBlanks(#base_form,#initial,#infix,#last_name)
& #rel := IsReliablePersonName(#full_name).
```

Another matching rule, this time for verbal structures, is given below. This is a simple rule for recognizing single verbs.

Example:

```
verb_detector :> (

                (gazetteer & [GTYPE: "verb", VOLITIONAL: #volition,

                             EPISTEMIC: #epistem, TEMPORAL: #temp,

                             SURFACE: #base_form])):verb
```

-> verb: verbal_analysis & [TYPE: "Verb", SURFACE:#base_form, VOLITIONAL: #volition, EPISTEMIC: #epistem, TEMPORAL: #temp].

Next sections explain the structure of the system in terms of grammar application strategy.

# CHAPTER 6

# 6. HOW THE SYSTEM WORKS

The analysis is done on text files. The algorithm processes that text line by line. Although most of the times a line is a paragraph, according to source text, it can be just one sentence.

After splitting lines into sentences, the system applies the grammar on the basis of the lexicon, in order to identify TAM value bearing structures and noun phrases, which can be person or group mentions.

The output of the system consists of all detected structures with their feature names, feature types and feature values. Although the output can be limited by setting ExPRESS parameters, it is given without eliminating anything. A filtering of the output can be done according to TAM categories easily. The structure of the system is depicted in Figure 1.

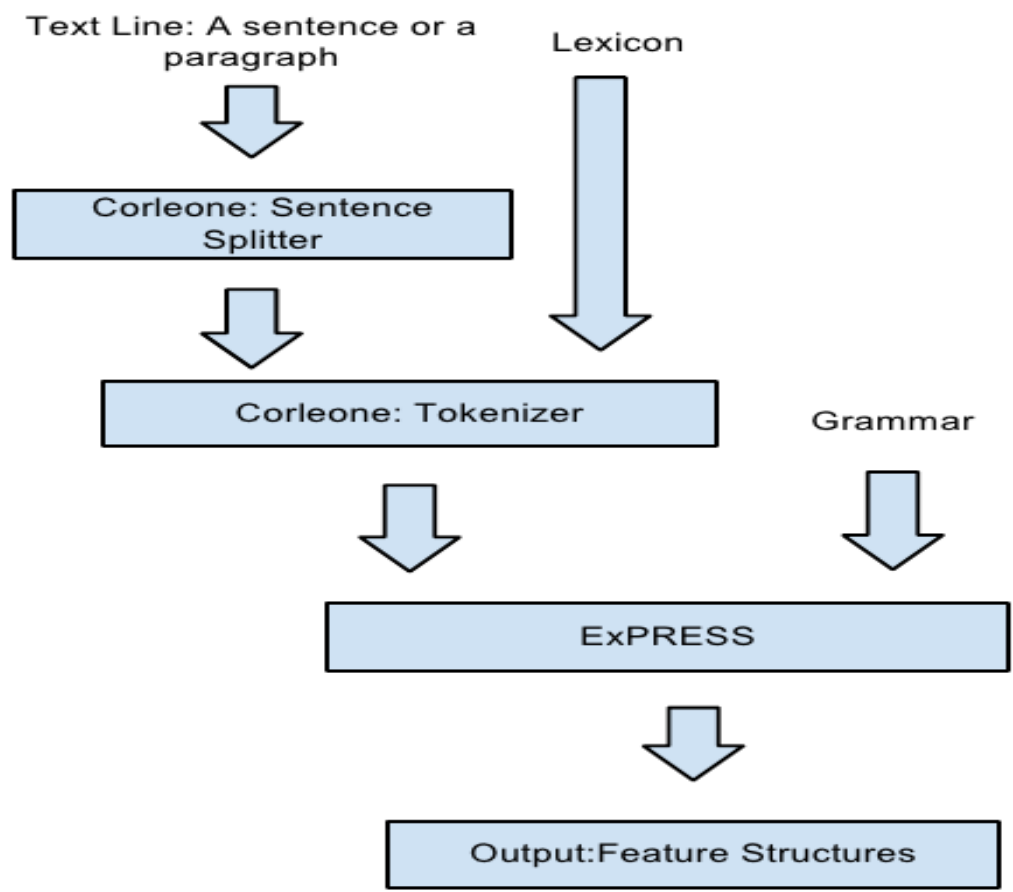

**Figure 1.** System structure.

The Express grammar can be written in several levels. This feature allows to concentrate on different aspects of texts and do shallow parsing, by using clues incrementally. The structure and number of the levels can be arranged according to the needs of the task.

The current system's first level grammar detects noun phrases of named entities and group names as multi-word expressions. Adverbs and verbs are also detected. The identified structures are transferred to subsequent grammar levels as feature structures.

The intermediate level of the grammar takes as input the output of the first level, and discovers the relation between these structures in order to build more comprehensive feature structures. It identifies relations between detected persons, person groups, verbal and adverbial clauses. In case of NPs, the relations can be characterized by commas or connectives like 've' (*and*) or 'ile' (*with*).

The second level grammar searches for matching types by considering intermediate level grammar's results. The main function of this level is to search for adverbials around recognized structures one more time. This level also contains grammar rules for identification of sentence feature structure.

## *6.1 Sample Analyses*

For the purpose of exposition, two articles from the Van earthquake topic were given as input to the system. Articles can be found in Appendix C. Some outputs from selected analyses are explained below. It will be noted that the same input sentence may be mapped to more than one output, as a result of matching by different lexical entries or different grammar rules.

1. Line : Kızılay: "İçme suyu ve iş makinalarına ihtiyaç var"
   Line (English): Turkish Red Crescent: "There is need for drinking water and for construction machinery. "
   a) İçme suyu ve iş makinalarına ihtiyaç var
   sentence_analysis
   TYPE : Sentence
   SURFACE : --içme suyu-iş makinası ihtiyaç var
   VOLITIONAL : asserted ( )
   EPISTEMIC : certain ( \ \ )
   TEMPORAL : perfect ( )

VERB : ihtiyaç var

HUMANITARIAN_NEED : --içme suyu-iş makinası

RULE : request_sentence

b) İçme suyu ve iş makinalarına ihtiyaç var

sentence_analysis

TYPE : Sentence

SURFACE : --içme suyu-iş makinası ihtiyaç var

VOLITIONAL : asserted ( )

EPISTEMIC : certain ( \ \ )

TEMPORAL : simultaneous ( )

VERB : ihtiyaç var

HUMANITARIAN_NEED : --içme suyu-iş makinası

RULE : request_sentence

As can be seen from the "RULE" feature, the 'request sentence' grammar rule operates on the sentence. From this rule, a request verb and some humanitarian need, as given in the "HUMANITARIAN_NEED" feature, items are present. Since the TAM category attributes of the verb 'var' have two different values (PERFECT and SIMULTANEOUS) in the lexicon, the output produces two different results for each variation. The TAM class dimension values are given in "VOLITIONAL", "EPISTEMIC" and "TEMPORAL" features. Their values are assigned by relevant rule by using lexicon entry of the sentence verb and other identified elements, which are not present in this sample.

2. Line: İlk olarak Erçiş'te büyük hasar olduğu 25 apartmanın ve bir öğrenci yurdunun çöktüğü tespit edildi.
   English: Initially, it is determined that there is extensive damage in Erciş, and that 25 apartments and a student dormitory have been collapsed.

a) 25 apartmanın ve bir öğrenci yurdunun çöktüğü tespit edildi

sentence_analysis

TYPE : Sentence

SURFACE : çöktüğü tespit edildi

VOLITIONAL : asserted ( )

EPISTEMIC : inferred ( \ \ )

TEMPORAL : anterior_ana_occ ( )

VERB : çöktüğü tespit edildi

RULE : subordination_sentence

b) 25 apartmanın ve bir öğrenci yurdunun çöktüğü tespit edildi

sentence_analysis

TYPE : Sentence

SURFACE : çöktüğü tespit edildi

VOLITIONAL : asserted ( )

EPISTEMIC : inferred ( \ \ )

TEMPORAL : perfect ( )

VERB : çöktüğü tespit edildi

RULE : subordination_sentence

Evidential verbs like "tespit edildi" are analyzed by subordination rules. The epistemic values of evidential verbs can be either INFERRED or CONJECTURED, which is the former in the above example. Although this verb (and other evidential verbs) do not have any specific temporal value in the lexicon, this value is taken from the factive subordinated verb 'çöktüğü' (*that ... collapsed*) which denotes a disastrous event. Two different results for TAM values again come from this verb forms two different possible TAM classes entered to the lexicon.

3. Line: Van Belediye Başkanı Bekir KAYA: "Şu anda kent merkezinde 3 noktada göçükler meydana geldi.

English: Mayor of Van city, Bekir KAYA: "By now, there are three sites with collapsed buildings in the city center."

a) göçükler meydana geldi

sentence_analysis

TYPE : Sentence

SURFACE : göçükler meydana geldi

VOLITIONAL : asserted ( )

EPISTEMIC : certain ( \ \ )

TEMPORAL : anterior ( )

VERB : meydana geldi

DISASTER : göçükler

RULE : incidence_sentence

b) göçükler meydana geldi

sentence_analysis

TYPE : Sentence

SURFACE : göçükler meydana geldi

VOLITIONAL : asserted ( )

EPISTEMIC : certain ( \ \ )

TEMPORAL : anterior_ana_occ ( )

VERB : meydana geldi

DISASTER : göçükler

RULE : incidence_sentence

c) göçükler meydana geldi

sentence_analysis

TYPE : Sentence

SURFACE : göçükler meydana geldi

VOLITIONAL : asserted ( )

EPISTEMIC : certain ( \ \ )

TEMPORAL : perfect ( )

VERB : meydana geldi

DISASTER : göçükler

RULE : incidence_sentence

All of the above analyses were produced by the incidence verb rule. The three distinct values for the temporal category reflect the fact that -DI can be used for deictic/simple past (ANTERIOR), sequential/anaphoric past (ANTERIOR_ANA) as well as for varieties of PERFECT. The incidence sentence rule takes into account incidence verbs, any occurring adverbials and disaster names to create the final output. Therefore, TAM class dimension values are taken from the combination of the verb and the adverbials and the "DISASTER" feature is taken from the disaster names that occurs just before (or around) of these verbal and adverbial structures. Although disaster names can be taken automatically as anything that occurs just before the incidence verbs, currently it is allowed to match just disaster names that are present in the lexicon. The aim of this approach is to control the output.

4. Line: Yarın yağmur, çarşamba günü de kar yağışı öngörülürken, perşembe günkü yağışların ardından bölgenin cuma günü ve hafta sonu yağışsız ve nispeten daha sıcak olması bekleniyor.

English: While tomorrow is forecast to be rainy and Wednesday snowy, it is expected that after the precipitations on Thursday, there will be no precipitation on Friday and at the weekend, and that it will get warmer.

a) sonu yağışsız ve nispeten daha sıcak olması bekleniyor

sentence_analysis

TYPE : Sentence

SURFACE : olması bekleniyor

VOLITIONAL : asserted ( )

EPISTEMIC : conjectured ( \ \ )

TEMPORAL : posterior ( )

VERB : olması bekleniyor

RULE : subordination_sentence

b) sonu yağışsız ve nispeten daha sıcak olması bekleniyor

sentence_analysis

TYPE : Sentence

SURFACE : olması bekleniyor

VOLITIONAL : asserted ( )

EPISTEMIC : conjectured ( \ \ )

TEMPORAL : prospective ( )

VERB : olması bekleniyor

RULE : subordination_sentence

The evidential verb from 'bekleniyor' is taken as signaling CONJECTURED rather INFERRED, since the inference involved is not definitive than 'olması' is analyzed as a future subordination verb form, which has two possible temporal values; namely, PROSPECTIVE and POSTERIOR. The system could capture what is conjectured as well. However, subordination sentence rules only focus on TAM class combinations.

5. Line: Ziyaret sırasında öğrencilerin 3 günlük tatilin yetmeyeceğini söylemesi üzerine Dinçer, Van'da okulların bir hafta tatil edildiğini söyledi.

   English: During the visit, as students said that a three-days break will not enough, Mr. Dinçer said that schools have been suspended for one week.

a) Dinçer, Van'da okulların bir hafta tatil edildiğini söyledi

sentence_analysis

TYPE : Sentence

SURFACE : Dinçer edildiğini söyledi

VOLITIONAL : hearsay ( )

EPISTEMIC : ( \ \ )

TEMPORAL : anterior_ana_occ ( )

PEOPLE : Dinçer

VERB : edildiğini söyledi

RULE : subordination_sentence

b) Dinçer, Van'da okulların bir hafta tatil edildiğini söyledi

sentence_analysis

TYPE : Sentence

SURFACE : Dinçer edildiğini söyledi

VOLITIONAL : hearsay ( )

EPISTEMIC : ( \ \ )

TEMPORAL : perfect ( )

PEOPLE : Dinçer

VERB : edildiğini söyledi

RULE : subordination_sentence

c) Dinçer, Van'da okulların bir hafta tatil edildiğini söyledi

sentence_analysis

TYPE : Sentence

SURFACE : Dinçer edildiğini söyledi

VOLITIONAL : hearsay ( )

EPISTEMIC : ( \ \ )

TEMPORAL : simultaneous ( )
        PEOPLE : Dinçer
        VERB : edildiğini söyledi
        RULE : subordination_sentence

Another example with sample is presented here. The combination of subordinator verb "edildiğini" and the relative clause verb resulted in the given feature structures. Since none of these verbs possess any epistemic dimension, this part in the sentence analysis remains null. The subordinator verb "edildiğini" has three different values for its temporal dimension. Therefore the system has an output feature structure for each of these entries. Moreover, the "PEOPLE" feature is extracted, as a person name was identified in the sentence.

# CHAPTER 7

# 7. AVENUES FOR FURTHER RESEARCH AND DEVELOPMENT

## 7.1. Discourse Considerations, Pragmatic Parameters, and Other Possible Improvements in coverage

The developed system aims to analyze sentences, and the interpretation of sentences is known to be discourse-sensitive. The effect of discourse on the interpretation of the sentences involve not only obvious phenomena like pronoun resolution, but also so-called anaphoric tenses and moods. Therefore, taking advantage of discursive clues in a paragraph-based approach would be a further step for the current project. Identification of discourse modes (Smith, 2004), as well as detection and proper analysis of discourse connectives and punctuation may help disambiguate TAM-category assignments.

For example, in the Narrative discourse mode, sentences are typically past perfective and would be semantically analyzed as expressing anaphoric 'sequential past' (ANTERIOR_ana_occ), rather then simple deictic past (ANTERIOR). However, in the Report mode, there will be more variation in tense, but tenses will be interpreted in a deictic way (i.e., relative to the time of the report). The present perfect, present progressive, and the future markers will all express direct relevance to current situation. In the Description mode, one is more likely to encounter stative predicates, continuous or progressive aspect and to some degree, habitual and generic statements. The aspect-temporal values indicating these (ongoingness) will typically be relative some past reference time, or the time of report.

In the news reports genre, the most widely encountered discourse modes are Narrative, Report and Description. As such, sequential past, deictic temporal values (perfect, progressive, prospective) are common, and deviations from epistemic certainty and assertive mode are rare with the exception of reported speech. One will not expect many atemporal-generic expressions, nor elaborate epistemic strategies (possibility, counter-factuality, conditionality) which may dominate the Information and Argument modes.

The CORLEONE and ExPRESS systems can be configured for analyses at several levels. These levels can be used for incremental analysis of any sort. As a prospect, it is possible to implement a third cascaded grammar level: A discourse grammar. The improvement of event extraction systems by including discourse-level analyses was demonstrated in sentiment analysis studies (Ferguson et al., 2009; Taboada, Brooke, & Stede, 2009). Discourse segmentation studies (Informatique, Torres, & Informatique, 2010) can be used in such an improvement.

The rules in this grammar level may profit from discourse connectives in assigning TAM values to sentences or clauses. Orthographic clues, including punctuation can be used for splitting complex sentences which include multiple propositions. Indeed, punctuation has been shown to be useful for comprehensive analyses of texts (Say & Akman, 1996, 1997).

This level can also use some heuristics to detect the discourse mode of the paragraph. The output structure for the discourse grammar could be similar to sentence feature structure. It can additionally contain information on discourse mode as well as statistical information about the special elements (e.g., these denoting humanitarian needs, disasters, and requests in a crisis menagament system).

A full analysis of TAM markers cannot be accomplished without taking into account contextual (deictic) parameters like identity of the reporter, reporting (speech) time, reference time (temporal center), epistemic center and volitional center. Holding a separate register for these pragmatic parameters (e.g., Giunchiglia & Bouquet, 1997) could be an effective strategy. Online news reports analysis can offer some advantages of simplicity, so that report time can be taken as the reference time, and speaker as the reporter. Moreover, it has been shown that the use of deixis in newspaper articles is limited (Ehlich, 1989).

A comprehensive study of time expressions in Turkish may enable the system to be more temporal aware, and may result in more precise analyses.

As to the use of the system in crisis management, event type specific categorizations can be elaborated. For instance, different natural disasters can be identified as, e.g., earthquakes, floods, storms etc. Important verbal forms, which can indicate the severity of events, can be detected, such as 'injured', 'died' etc.

## 7.2 Possible system improvements

Instead of manual creation of the lexicon files, a machine learning algorithm can be used for inferring TAM-categories for verb forms, on the basis of their morphological structure. Although such a system will require human annotators for feedbacks and for annotating sentences, the current system can be used to pre-annotate the data.

As to the creation of the corpus, the current manual approach helped us control the parameters of the system and understand the applicability of the method. The gathering and clustering of news articles too can actually be done automatically for large amount of articles (Jayabharathy, Kanmani, & Parveen, 2011; Krishna, 2010; Mokris & Skovajsova, 2008). In addition to mainly computer science-based methods that dominate document and text clustering, semantics- and cognitive science-based algorithms (Choudhary & Bhattacharyya, 2002; Guo, Shao, & Hua, 2009; Hotho, Maedche, & Staab, 2001; Reforgiato Recupero, 2007) are also under development.

Although the system recognizes cases of NPs, it does not distinguish between the subject and the object(s) in a sentence. Since the lexicon has different forms of group names and apostrophe interpretation for proper names for case markers, this step is not hard to implement. Such an extension would enable the system to have a more precise output which is closer to a first order logic-based format.

# CHAPTER 8

# 8. CONCLUSION

The current thesis project developed an event extraction software system which focuses on tense, aspect and mood values of sentences, in the context crisis management. To this end, a small representative corpus was assembled, a lexicon file was prepared in the standard of CORLEONE, and a cascaded-grammar rules were implemented on the basis the extraction pattern engine ExPRESS. TAM analyses were guided by the theory of anchoring relations developed in Temürcü (2007, 2011).

The output of the system consists in feature structures which include temporal, epistemic and volitional/illocutionary dimensions of each verb and adverb, and derivatively, for each sentence. This representation provides a simplified grammatical structure, which includes parts of a sentence relevant for TAM analyses as well as those useful for cirisis management. This structure is convertable to a first order logic formula, or to a Discorse Representation Structure (Kamp & Reyle, 1993). After relevant improvements, the system has the potential to be used as an interface for multi-modal reasoning and inference tasks.

# APPENDICES

## A. SOME FREQUENT TAM COMBINATIONS IN NEWS

This section includes some of the frequently occurring TAM combinations in news genre, along with examples which show possible TAM values for these combinations. The analyses and most of the examples are taken from Temürcü (2007).

**1. V-Xr**

**A.** ASSERTED + GENERAL FACT + ATEMPORAL

(i)      Kedi-ler karanlık-ta gör-ür.

        'Cats see in the dark.'

(ii)     Mustafa hayvan-lar-ı çok sev-er.

        'Mustafa loves animals so much.'

**B.** ASSERTED + GENERAL FACT + PERFECT/persisting/

(i)      Sekiz yaş-ım-dan beri saz çal-ar-ım.

        'I have been playing saz since I was eight.'

**C.** ASSERTED + PROBABLE + POSTERIOR

(i)      Ali bu gece bir ihtimal-le / belki / herhalde gel-**ir**.

        'Ali might / may / will probably come tonight.'

**D.** ASSERTED + PROBABLE + PROSPECTIVE

(i)      Ali birazdan gel-ir.

        'Ali will (probably) come in a minute.'

## 2. V-Xr-(y)DI

**A**. ASSERTED+CERTAIN+ANTERIOR$_{ANA}$(ATEMPORAL)

(i)      Ali eskiden çok sigara iç-er-di.

        'Ali used to smoke too much.'

B. ASSERTED+ ¬CERTAIN$_{ANA}$(PROBABLE) + POSTERIOR

(i)      Ahmet bura-da ol-sa-ydı pikniğ-e gid-er-**di**-k.

        'If Ahmet was here we would go out for a picnic.'

## 3. V-Iyor

**A.** ASSERTED + CERTAIN + SIMULTANEOUS

(i)      Ahmet (şimdi / şu an-da) kahvaltı yap-**ıyor**.

        'Ahmet is has having his breakfast (now / at the moment).'

**B.** ASSERTED + CERTAIN + RECURRENT

(i)      Hasan (arada bir/sık sık) uğr-uyor.

        'Hasan drops by occasionally/frequently.'

**C.** ASSERTED + CERTAIN + PERFECT/persisting/

(i)    Osman beş yıl-dır bu muhit-te otur-**uyor**.

       'Osman has been living in this district for five years.'

**D.** ASSERTED + CERTAIN + ATEMPORAL

(i)    Ay döngü-sü-nü 28 gün-de tamaml-**ıyor.**
       'The moon completes its cycle 28 days.'

**E.**  ASSERTED + NEW INFORMATION + POSTERIOR

(i)    Kayıt-lar 12 Haziran'da başl-ıyor

       'The registration period starts on June 12.'

**F.**  ASSERTED + NEW INFORMATION + PROSPECTIVE/imminent/

(i)    Kayıt-lar başl-ıyor

       'The registration period is about to start'

## 4. V-Iyor-(y)DI

**A.** ASSERTED + CERTAIN + ANTERIOR_ANA(SIMULTANEOUS)

(i)    Gel-diğ-im-de Ahmet yemek y-iyor-du.
       'When I came Ahmet was having his meal.'

## 5. V-mAkta

**A.** ASSERTED + CERTAIN + SIMULTANEOUS

(i)    Ekolojik denge hızla bozul-makta.
       'The ecological balance is being destroyed rapidly.'

**B**. ASSERTED + CERTAIN + RECURRENT

(i)    Bakanlık tesis-ler-i düzenli olarak denetle-mekte.

'The ministry is regularly inspecting the establishments.'

**C**. ASSERTED + CERTAIN + PERFECT/persisting/

(i)    Şirket 15 yıl-dır bilişim sektör-ü-nde faaliyet göster-**mekte**.

'The company has been active in the informatics sector for 15 years.'

## 6. V-mAktA-DIr

**A**. ASSERTED + GENERAL FACT + SIMULTANEOUS

(i)    Ekolojik denge hızla bozul-makta-dır.

'The ecological balance is being destroyed rapidly.'

**B**. GENERAL STATEMENT + CERTAIN + SIMULTANEOUS

(i)    İlgili birimlerimiz konu üzerinde çalışmaktadır.

**C**. GENERAL STATEMENT + CERTAIN + RECURRENT

(i)    Bakanlık tesis-ler-i düzenli olarak denetle-mekte-dir.

'The ministry is regularly inspecting the establishments.'

(ii)   Bölge-de çok sık deprem etkinliğ-i kayded-il-mekte-**dir**.

'Very frequent earthquake activities are being recorded in the region.'

**D**. GENERAL STATEMENT + CERTAIN + PERFECT/persisting/

(i)    Şirket 15 yıl-dır bilişim sektör-ü-nde faaliyet göster-**mekte**-dir.

'The company has been functioning in the informatics sector for 15 years.'

## 7. V-mAktA-(y)DI

**A.** ASSERTED + CERTAIN + ANTERIOR$_{ANA}$(SIMULTANEOUS)

(i)     Şirket 15 yıl-dır bilişim sektör-ü-nde faaliyet göster-**mekte-ydi**.

## 8. V-(y)AcAG

**A.** ASSERTED + CERTAIN + POSTERIOR

(i)     3 Aralık'ta kuralar çekilecek

**B.** ASSERTED + CERTAIN + PROSPECTIVE

(i)     Halk bölgeden tahliye edilecek

## 9. V-(y)AcAG-DIr

**A.** ASSERTED + GENERAL FACT + POSTERIOR

(i)     Petrol rezerv-ler-i azal-dığ-ı-nda bir enerji kriz-i yaşa-n-**acak-tır**.

**B.** GENERAL STATEMENT + CERTAIN + POSTERIOR

(i)     Toplantı sonra-sı-nda ayrıntılı açıklama yap-ıl-acak-tır.

'A detailed statement will be issued after the meeting.'

**C.** GENERAL STATEMENT + GENERAL FACT + POSTERIOR

(i)     Suçlu mutlaka yakala-n-acak-tır.

'The criminal is bound to be captured.'

## 10. V-(y)AcAG-(y)DI

**A.** ASSERTED + ¬CERTAIN$_{ANA}$(CERTAIN)+ ANTERIOR$_{ANA}$(POSTERIOR)

(i)    Beş-te önemli bir toplantı ol-a**cak**-tı.

**B.** ASSERTED + ¬CERTAIN$_{ANA}$(CERTAIN) + ANTERIOR$_{ANA}$(PROSPECTIVE)

(i)    Yetkililer bölgeyi ziyaret ed-ecek-ti

## 11. V-DI

**A.** ASSERTED + CERTAIN + ANTERIOR

(i)    Dün önemli bir toplantı yapıl-**dı**.

**B.** ASSERTED + CERTAIN + PERFECT

(i)    Yardım yerine ulaştı

**C.** ASSERTED + CERTAIN +  ANTERIORANA.OCC

(i)    { Dün bir toplantı yapıldı.} Afet merkezi kurulmasına karar veril-**di.**

## 12. V-mIş-DIr

**A.** GENERAL STATEMENT + CERTAIN + ANTERIOR

(i)    Bakanlar kurulu saat beşte toplanmıştır.

**B.** GENERAL STATEMENT + CERTAIN + PERFECT

(i)    Yardım yerine ulaşmıştır.

## 13. V-mIş-(y)DI

**A.** ASSERTED + CERTAIN + ANTERIOR$_{ANA}$(PERFECT)

(i)    Dün saat beşte yardım yerine ulaşmıştı.

**14. V-mIş ol-acak**

**A.** ASSERTED + CERTAIN + POSTERIOR<sub>ANA</sub>(PERFECT)

(i)      Yarın saat beşte yardım yerine ulaşmış olacak.

**15. V-(y)Abilir**

**A.** ASSERTED + PROBABLE + PROSPECTIVE

(i)      Bölgede artçı sarsıntılar olabilir.

**16. P-DIr** (*P* for a nominal predicate)

**A.** ASSERTED + GENERAL FACT + ATEMPORAL

(i)      Kediler etçil-dir

**B.** GENERAL STATEMENT + CERTAIN + SIMULTANEOUS

(i)      Alana girmek tehlikeli-dir

**C.** GENERAL STATEMENT + CERTAIN + PERFECT/persisting/

(i)      Başkan on yıl-dır bu bölüm-de görevli-dir.

**17. P-(y)DI** (*P* for a nominal predicate)

**A.** ASSERTED + CERTAIN + ANTERIOR<sub>ANA</sub>(SIMULTANEOUS)

(i)      O sırada evde yalnız-**dı**

**B.** ASSERTED + CERTAIN + ANTERIOR<sub>ANA</sub>(PERFECT/PERSISTING/)

(ii)     On yıldır işsiz-**di**

# B. VAN EARTHQUAKE ARTICLES

**Article 1**

http://gundem.milliyet.com.tr/kizilay-icme-suyu-ve-is-makinalarina-ihtiyac-
var-/gundem/gundemdetay/23.10.2011/1454214/default.htm , 23.10.2011

Kızılay: "İçme suyu ve iş makinalarına ihtiyaç var"

Van'da meydana gelen depremin ardından Türk Kızılay'ı alarma geçti. Kızılay,
"Bölgede arama kurtarma ekipleri, iş makinaları ve içme suyu ihtiyacı bulunuyor"
dedi

Van Tavanlı merkezi 6.6 büyüklüğündeki depremin ardından Türk Kızılay'ı alarma
geçti. Konuya ilişkin yazılı bir açıklama yapan Kızılay, "Ankara'daki Türk Kızılayı
Afet Operasyon Merkezi'nde bir kriz masası oluşturulduğu"nu açıkladı.

Kızılay'ın Erzurum, Muş ve Elazığ Yerel Afet Müdahale ve Lojistik Merkezlerinden
çok sayıda malzeme gönderildiği bildirilen açıklama şöyle:

"Yerel lojistik merkezlerinden çok sayıda çadır, battaniye ve gıda malzemesi bölgeye
doğru yola çıkarıldı. Türk Kızılayı'nın ekipleri bölgede durum ve ihtiyaç tespit
çalışmalarına devam ediyor. Bölgedeki kan merkezleri de yaralılar için gerekli kanı
ulaştırmak için çalışmalarına başladı.

İlk olarak Erçiş'te büyük hasar olduğu 25 apartmanın ve bir öğrenci yurdunun çöktüğü tespit edildi. Kızılay'ın yerel ekipleri çöken öğrenci yurdundan yaralıları çok sayıda yaralı çıkardı. Bölgede arama kurtarma ekipleri, iş makinaları ve içme suyu ihtiyacı bulunuyor. Türk Kızılayı Genel Müdürü Ömer Taşlı da bölgeye doğru yola çıktı."

http://www.samanyoluhaber.com/h_710613_Gundem-kizilay-acikladi-ihtiyac-duyulan-malzemeler.html, 24.10.2011

En çok ihtiyaç duyulan malzemeler ,23.10.2011,22:03:42

Van'da meydana gelen depremin ardından Türk Kızılay'ı alarma geçti. Bölgednin özellikle battaniye, çadır ve içme suyu ihtiyacı bulunuyor. Bütün acil aramaların ise 112 ve 155 ile yapılacağı belirtildi.

Van Tavanlı merkezi 7.2 büyüklüğündeki depremin ardından Türk Kızılay'ı alarma geçti. Konuya ilişkin yazılı bir açıklama yapan Kızılay, "Ankara'daki Türk Kızılayı Afet Operasyon Merkezi'nde bir kriz masası oluşturulduğu"nu açıkladı.

Kızılay'ın Erzurum, Muş ve Elazığ Yerel Afet Müdahale ve Lojistik Merkezlerinden çok sayıda malzeme gönderildiği bildirilen açıklama şöyle:

"Yerel lojistik merkezlerinden çok sayıda çadır, battaniye ve gıda malzemesi bölgeye doğru yola çıkarıldı. Türk Kızılayı'nın ekipleri bölgede durum ve ihtiyaç tespit çalışmalarına devam ediyor. Bölgedeki kan merkezleri de yaralılar için gerekli kanı ulaştırmak için çalışmalarına başladı. Bölgede arama kurtarma ekipleri, iş makinaları ve içme suyu ihtiyacı bulunuyor."

Yardım malzemelerinin bölgeye aktarımı için tüm hazırlıkların tamamlandığı ifade edilen açıklamada, "Türk Kızılayı bölgedeki gelişmeleri yakından takip etmektedir" denildi.

## ÇADIR VE BATTANİYE İHTİYACI VAR

Kızılay, Van'da meydana gelen depremin ardından bölgedeki depolarından bölgeye çadır, battaniye ve diğer yardım malzemelerini göndermeye başladı.

Van'dan meydana gelen depremin ardından, Türk Kızılayı Genel Başkanı Ahmet Lütfi Akar'ın talimatıyla bütün birimler acil duruma geçti. Kızılay Afet Operasyon merkezinde gelişmeleri takip etmek için kriz masası oluşturuldu.

Erzurum başta olmak üzere yerel lojistik depolarından çadır, battaniye ve diğer ihtiyaç malzemeleri de Van'a gönderilmeye başlandı.

Bu arada Van'da bulunan bir ekip de ihtiyaç ve durum tespiti çalışmasını sürdürüyor, bu tespite göre yardımlar koordine edilecek.

## ACİL ARAMALAR 112 VE 155 İLE YAPILACAK

Van'da meydana gelen depremde enkaz altında kalan ve zor durumda olan depremzedelere yardım etmek için aşağıdaki özellikle acil aramalar 112 ve 155 ile yapılacağı belirtildi.

Ayrıca Kızılay, yardımda bulunmak isteyenlerin 168'i arayabileceğini de açıkladı.

## DEPREM BÖLGESİNE SEVK DEVAM EDİYOR

Başbakanlık Afet ve Acil Durum Yönetimi Başkanlığı (AFAD), 38 ilden yaklaşık bin

275 arama kurtarma personeli, 174 araç, 290 sağlık personeli, 43 ambulans, 6 hava ambulansının deprem bölgesine gönderildiğini bildirdi.

AFAD'dan yapılan yazılı açıklamada, depremin ardından ilgili tüm Bakanlık, kurum ve kuruluşlarla koordinasyon sağlandığını, ihtiyaçların temini ve tespiti için çalışmalara başlandığını belirtildi.

Arama kurtarma ve ilk yardım çalışmalarının kesintisiz devam ettiği bildirilen açıklamada, şunlar kaydedildi:

"Deprem bölgesine, 38 ilden yaklaşık bin 275 arama kurtarma personeli, 174 araç, 290 sağlık personeli, 43 ambulans, 6 hava ambulansı gönderilmiştir. Van'a 4 uçakla ekipler gönderilmiş olup 6 uçağın da arama kurtarma ve ilk yardım ekiplerini götürecek şekilde planlaması yapılmıştır.

Ayrıca GSM operatörlerine ait 1 mobil baz istasyonu kurulmuş, çevre illerden de 9 mobil baz istasyonu bölgeye sevk edilmiştir. Başkanlığımızdan 20 kişilik bir teknik heyet Van'a intikal etmiştir."

Açıklamaya göre kurumlardan sevk edilen ekip ve ekipmanlar şöyle:

"Azerbaycan Hükümeti: 120 kişilik arama kurtarma ekibi. Sağlık Bakanlığı: 2 ambulans uçak, 4 helikopter, sahra hastanesi. Çevre Bakanlığı: 3 arama kurtarma köpeği. Emniyet Genel Müdürlüğü: 88 personel. Erzurum İl Sağlık Müdürlüğü: UMKE ekibiyle 1 adet şişmehastaneçadırı, 1 hava ambulansı. Kızılay: 96 personel ile Muş, Elazığ ve Erzurum'dan çadır ve aş evi. 2 bin 571 çadır, 7 bin 546 battaniye, 200 soba, 7 seyyar mutfak, 284 mutfak seti, 1 fırın, bin 120 gıda paketi, bin 320 su, 5 jeneratör, 5 bin kumanya. Türk Hava Kurumu: 1 ambulans uçak. Gökçen Havacılık: 1 ambulans. Aras Elektrik: 2 jeneratör, 3 araç. Cumhuriyet Başsavcılığı Cezaevi

Müdürlüğü: 3 cezaevi nakil aracı.AnkaraBüyükşehir Belediyesi: 20 kişilik arama kurtarma ekibi, 3 arama kurtarma köpeği. Konya Büyükşehir Belediyesi: 7 personel, 1 araç. GEA: 20 arama kurtarma personeli. AKUT: 40 arama kurtarma personeli. Türksat: 2 personel.Deniz FeneriDerneği: 60 ev Erzurum'dan yola çıktı. Kimse Yok mu?: 1 kamyon battaniye. Ulusal Acil Durum Arama Kurtarma Derneği (NESAR): 10 kişilik arama kurtarma ekibi. Şanlıurfa DEDAŞ: 9 personel, 3 araç ve 2 jeneratör."

**Article 2**

http://www.haberturk.com/yasam/haber/681925-su-ekmek-ve-cadira-ihtiyacimiz-var-,24.10.2011

"Su, ekmek ve çadıra ihtiyacımız var!" , 23.10.2011, 19.44

Van Belediye Başkanı Bekir KAYA, yardım çağrısında bulundu...

Van Belediye Başkanı Bekir KAYA: "Şu anda kent merkezinde 3 noktada göçükler meydana geldi. Ancak bu panik ortamında çok fazla bilgi kirliliği var. Doğal olarak ilk deprem anında her kurum ne yapabilirim paniğiyle bir şeyler yapmaya çalıştı. Yardım ekipleri yolda ama hala pek ulaşan olmadı bölgeye. Ben tüm yardım kuruluşlarına şunu söylemek istiyorum. Lütfen hiç talep beklemeden herkes elinden gelen yardımı hemen ulaştırmaya baksın. Kimse Van'dan talep beklemesin. Herkes yardım için elinden geleni yapsın."

"Su, ekmek ve çadıra ihtiyacımız var!"
"En çabuk ulaşılacak yerler Erciş ve yakınındaki köylerdir. Şu anda çadırlara, battaniyeye, içme suyuna, temel gıda maddelerine, suya, ekmeğe çok ama çok acil ihtiyaç var. Kentin içinde yıkılmayan çoğu binada da oturmak, barınmak artık

mümkün gözükmüyor."

"Hasar görmüş yapılara kesinlikle girmeyin"
"Ben bütün vatandaşlarımızdan şunu rica ediyorum; Arama kurtarma ekiplerinin bulunduğu yerlede toplanmayın kalabalık oluşturmayın. Hasar görmüş yapılara kesinlikle girmeyin. Mümkün olduğunca herkes birbirine yardımcı olsun."

OKULLAR 1 HAFTA TATİL EDİLDİ
Milli Eğitim Bakanı Ömer Dinçer, Van'da eğitime 1 hafta ara verildiğini bildirdi. Dinçer, Başbakan Recep Tayyip Erdoğan'ın Van Devlet Hastanesine yaptığı ziyarete eşlik etti. Ziyaret sırasında öğrencilerin 3 günlük tatilin yetmeyeceğini söylemesi üzerine Dinçer, Van'da okulların bir hafta tatil edildiğini söyledi.

Van Yüzüncü Yıl Üniversitesi'nde de (YYÜ), kentte yaşanan 7.2'lik depremin ardından eğitime bir hafta ara verildi. YYÜ Rektörü Prof. Dr. Peyami Battal, sadece kampüs içinde 3 binada hafif çaplı hasar olduğunu, yaralanma veya can kaybı yaşanmadığını bildirdi.

Van Yüzüncü Yıl Üniversitesi Rektörü Prof. Dr. Peyami Battal, öğrencileri memleketlerine yönlendirme işlemlerini tamamladıklarını belirterek, 4 binin üzerinde öğrenciyi evlerine gönderdiklerini bildirdi.

Öğrencilerin memleketlerine yönlendirilmesi işlemlerinin akşam boyunca sürdüğünü belirten Battal, öğrencilerin bir kısmının uçaklarla, bir kısmının diğer araçlarla Van'dan ayrıldığını söyledi.

Erciş'teki öğrencilerin de araçlarla üniversite kampüsüne getirildiğini ve aynı şekilde memleketlerine yönlendirildiğini ifade eden Battal, hala kentte bulunan az sayıdaki öğrencinin ise daha önceden biletlerini aldıkları için yolculuk saatini beklediklerini aktardı.

Battal, şunları kaydetti:

"Öğrencilerimizi memleketlerine yönlendirme işlemleri tamamlandı. 4 binin üzerinde öğrenciyi evlerine gönderdik. Otobüslerin olmadığı noktalarda çevre illerden üniversite ve firmaların araçlarını yönlendirdik. Desteklerinden dolayı çevre üniversitelerin rektörlerine, acentelere teşekkür ediyoruz. Öğrencilerin evlerine ulaşmaları noktasında Diyarbakır, Mersin, Erzurum, Iğdır gibi belli güzergahlar belirledik. Öğrencilerimiz en kısa sürede ailelerine ulaşacaklar."

Eğitime verilen 1 haftalık aranın gerekirse uzatılacağını ifade eden Battal, öğrencilerin, üniversitenin internet sitesinden durumu takip edebileceğini belirtti. Battal, personele idari izin verme konusunu da görüşeceklerini bildirdi.

Öte yandan yaralıların bir kısmının kurulan çadırlarda, bir kısmının da yeni faaliyete giren Bölge Araştırma Hastanesinde bulunduğunu anlatan Battal, ellerinden gelen bütün gayreti gösterdiklerini söyledi.

Bitlis'te de okullar 1 gün tatil edildi.

BÖLGEDE KAR YAĞIŞI BEKLENİYOR

Deprem bölgesindeki bir diğer sorun da havanın soğukluğu. Şu an 0 derecelere yakın bir sıcaklık söz konusuyken, bugün yağış beklenmiyor.

Yarın yağmur, çarşamba günü de kar yağışı öngörülürken, perşembe günkü yağışların ardından bölgenin cuma günü ve hafta sonu yağışsız ve nispeten daha sıcak olması bekleniyor.

Van'da meydana gelen depremde enkaz altında kalan ve zor durumda olan depremzedelere yardım etmek için aşağıdaki iletişim numaralarını kullanabilirsiniz:

Best Van Tur (444 00 65) ve Van Gölü Turizm (444 65 65) telefon numaraları ile irtibat kurup, kıyafet ve yiyecek yollanabiliyor.

Erciş Sosyal Yardımlaşma ve Dayanışma Vakfı: 0-432-351-59-06

Kızılay Yardım Hattı: 0 312-245-45-00, 0 312-430-18-14.

Kızılay, yardımda bulunmak isteyenlerin 168'i arayabileceğini de açıkladı.

AKUT: 2930'a AKUT yazan bir SMS mesaj ile AKUT'a 5TL katkıda bulunabilirsiniz.

Şişli Belediyesi (0212 288 75 76) Mavi Masa ile yarın sabah bir yardım daha çıkaracak.

**TEZ FOTOKOPİ İZİN FORMU**

<u>**ENSTİTÜ**</u>

Fen Bilimleri Enstitüsü ☐

Sosyal Bilimler Enstitüsü ☐

Uygulamalı Matematik Enstitüsü ☐

Enformatik Enstitüsü ☐

Deniz Bilimleri Enstitüsü ☐

<u>**YAZARIN**</u>

SOYADI    : HÜRRİYETOĞLU

ADI        : ALİ

BÖLÜMÜ   : Bilişsel Bilimler

TEZİN ADI : TENSE, ASPECT AND MOOD BASED EVENT EXTRACTION FOR SITUATION ANALYSIS AND CRISIS MANAGEMENT

TEZİN TÜRÜ: Yüksek Lisans ☐        Doktora ☐

1. Tezimin tamamı dünya çapında erişime açılsın ve kaynak gösterilmek şartıyla tezimin bir kısmı veya tamamının fotokopisi alınsın.
2. Tezimin tamamı yalnızca Orta Doğu Teknik Üniversitesi kullanıcılarının erişimine açılsın.(Bu seçenekle tezinizin fotokopisi ya da elektronik kopyası Kütüphane aracılığı ile ODTÜ dışına dağıtılmayacaktır.)
3. Tezim bir (1) yıl süreyle erişime kapalı olsun.(Bu seçenekle tezinizin fotokopisi ya da elektronik kopyası Kütüphane aracılığı ile ODTÜ dışına dağıtılmayacaktır.)

Yazarın imzası  …………………………        Tarih  …………………………